\documentclass[acmlarge]{acmart}
\usepackage[ruled,vlined]{algorithm2e}
\SetKwInput{kwStep}{Step 1}
\SetKwInput{kwStepp}{Step 2}
\usepackage{multirow}
\usepackage{graphicx}
\usepackage{enumitem}
\usepackage{titlesec}
\usepackage{xcolor,colortbl}
\usepackage{hyperref}

\definecolor{Gray}{gray}{0.85}
\definecolor{LightCyan}{rgb}{0.88,1,1}

\newcolumntype{a}{>{\columncolor{Gray}}c}
\newcolumntype{b}{>{\columncolor{white}}c}
\AtBeginDocument{%
  \providecommand\BibTeX{{%
    \normalfont B\kern-0.5em{\scshape i\kern-0.25em b}\kern-0.8em\TeX}}}

\setcopyright{acmcopyright}
\copyrightyear{2018}
\acmYear{2018}
\acmDOI{10.1145/1122445.1122456}

\acmJournal{POMACS}
\acmVolume{37}
\acmNumber{4}
\acmArticle{111}
\acmMonth{8}

\begin{document}

\title{Graph Neural Networks in IoT: A Survey}

\author{Guimin Dong, Mingyue Tang, Zhiyuan Wang, Jiechao Gao, Sikun Guo, Lihua Cai, Robert Gutierrez, Bradford Campbell, Laura E. Barnes, Mehdi Boukhechba} 
\affiliation{
  \institution{University of Virginia}
   \city{Charlottesville}
  \state{VA}
  \country{USA}
}
\email{emails:  {gd5ss, utd8hj, vmf9pr, jg5ycn, qkm6sq, lc3cp, rjg7ra, bradjc, lb3dp, mob3f}@virginia.edu}

\renewcommand{\shortauthors}{Dong, et al.}

\begin{abstract}
The Internet of Things (IoT) boom has revolutionized almost every corner of people's daily lives: healthcare, home, transportation, manufacturing, supply chain, and so on. With the recent development of sensor and communication technologies, IoT devices including smart wearables, cameras, smartwatches, and autonomous vehicles can accurately measure and perceive their surrounding environment. Continuous sensing generates massive amounts of data and presents challenges for machine learning. Deep learning models (e.g., convolution neural networks and recurrent neural networks) have been extensively employed in solving IoT tasks by learning patterns from multi-modal sensory data. Graph Neural Networks (GNNs), an emerging and fast-growing family of neural network models, can capture complex interactions within sensor topology and have been demonstrated to achieve state-of-the-art results in numerous IoT learning tasks. In this survey, we present a comprehensive review of recent advances in the application of GNNs to the IoT field, including a deep dive analysis of GNN design in various IoT sensing environments, an overarching list of public data and source code from the collected publications, and future research directions. To keep track of newly published works, we collect representative papers and their open-source implementations and create a Github repository at \href{https://github.com/GuiminDong/GNN4IoT}{GNN4IoT}.

\end{abstract}



\maketitle

\section{Introduction}\label{introduction}

Internet of things (or using its acronym IoT) refers to a set or sets of closely connected devices that form a network through wireless or wired communication technology and work collaboratively to achieve common goals for their users. It has been an ongoing development for the past few decades due to continuing advancements in sensors, processing and computing powers, embedded systems, and various communication protocols such as WiFi, Bluetooth, and RFID. There are several major components in today's IoT devices, including the sensing units, the processing units, the communication units, and its mechanical hardware components, that serve as the backbone of the devices and oftentimes with a user interface for both in-person and remote controls. Among these units, their sensing and communication capabilities are what distinguish them from non-IoT devices. With sensing capabilities, IoT devices can now perceive their positioned environment and understand the different environmental states as the surrounding contexts change. With communication capabilities, they are no longer working alone, but collaboratively with other connected devices and their users to achieve more complex tasks that were never thought possible on their own.

IoT systems are commonplace as they are embedded ubiquitously in most of our living and working spaces. Wearable devices, such as virtual glasses, wrist bands and smartwatches, and digital rings, are all collecting data about our surroundings and physiological states to keep us safer and healthier~\cite{garcia2018mental,boukhechba2018predicting}. They can be in ICU smart systems that consist of infrared cameras and smart beds to observe patients' vital signs such as blood pressure and body temperature~\cite{davoudi2019intelligent} or measure patient mobility~\cite{ma2017measuring}. They can be a network of surveillance cameras that monitor the community and trigger alerts when abnormal activities are detected~\cite{huang2021abnormal}. In addition, they can be applied in traffic monitoring using drivers' smartphone satellite/GPS signals to provide real time navigation services and traffic forecasts~\cite{jiang2021graph}; smart agriculture which leverages moisture and temperature sensors to monitor soil conditions to ultimately improve crop production~\cite{khoa2019smart,mekala2017survey}; smart grids and energy management to predict energy demands and allocate resources so that the generation of electricity from different plants can be scheduled intelligently and efficiently~\cite{li2017everything}. Most of us are familiar with the IoT systems that control our home and work spaces, whether it's lights, doors, speakers, appliances, etc., all of which enhance our quality of life~\cite{marikyan2019systematic}. The applications listed above provide a small glimpse into the realm of possibilities for what IoT sensing is capable of achieving.

The power of IoT sensing is realized through taking advantage of the collected data from the equipped sensors in different systems. 
Early IoT systems that consisted of homogeneous device types generated consistent data modality that is conducive to traditional data analytic approaches such as classical machine learning algorithms and other shallow learning methods~\cite{xie2018survey,cui2018survey}. Most shallow methods like tree-based models, SVM, and regression, require the raw sensing data to be preprocessed prior to generating predefined features for modeling. However, as more complex and heterogeneous IoT systems are being adopted in different application domains, there is also a rise in challenges. On one hand, handcrafted feature engineering requires significant domain knowledge from the corresponding field, which is both inefficient and limited in generalizability. On the other hand, such features are shallow and usually cannot encode complex interrelationships among the device activities, both spatially and temporally. Furthermore, data generated by a mixture of different sensors could be complementary to each other and/or missing in some scenarios which brings a fusion challenge when combining audio, video, motion and ambient data,  that are sampled at different frequencies and in multiple formats. To alleviate these issues, deep learning becomes increasingly adopted in processing IoT sensing data.



The development of deep learning methods in IoT sensing  have emerged as their adoption has grown. In computer vision based IoT systems, convolutional neural networks (CNNs) have played a central role due to their ability to abstract deep concepts in images~\cite{khan2018guide}. Various variants of (CNNs) have also been proposed to model IoT sensing data. For example, Vimal et al. applied CNNs in IoT based smart health monitoring systems~\cite{vimal2021iot}, while Li et al. proposed a CNN algorithm for electrical load forecasting in smart grids by converting all sensor data into image format~\cite{li2017everything}.
To encode temporal relationships among input instances, CNNs have also been combined with long-short-term-memory (LSTM) networks. LSTMs are a type of recurrent neural network (RNN) algorithm that boosts the forgetfulness problem in sequential inputs. For example, Hasan et al. presented a CNN-LSTM based approach for electricity theft detection in smart grid systems~\cite{hasan2019electricity}; Han et al. designed a hybrid CNN-LSTM framework for air quality modeling in metropolitan cities~\cite{han2021deep}; while Yang et al. applied a parallel convolutional RNN for emotion recognition using multi-channel EEG data~\cite{yang2018emotion}.


A more recent development of deep learning methods in IoT sensing focuses on graph neural network (GNN) and its variants. There are several benefits of applying a GNN to model IoT sensing data, besides what is provided by CNN and RNN. Indeed, both CNN and RNN can be treated as a simpler GNN with fixed-size grid graphs for CNN and line graphs for RNN, while typical GNN comes with more complex graphs, without a fixed form, with a variable size of unordered nodes, and a variable amount of neighbors for each node~\cite{zhou2020graph}.
The extra complexity in GNN empowers us to effectively encode complex relationships and interdependencies among devices in IoT sensing systems, and among data instances over time from IoT devices.
A wide range of applications in IoT sensing have leveraged GNN to represent their learning problems on the associated data, including works in team collaboration~\cite{wang2020mixed}, traffic monitoring and forecasting~\cite{cui2019traffic,jiang2021graph,zhang2020multi,chen2019gated,li2020graph}, remote scene classification~\cite{li2020multi}, robotic grasping~\cite{garcia2019tactilegcn}, movement analysis and object detection~\cite{jalata2021movement,shi2020point}, public service~\cite{zhang2020semi}, energy management~\cite{owerko2018predicting}, autonomous driving~\cite{li2020attentional,cai2020dignet}, social sensing~\cite{casas2020spagnn}, personal health~\cite{dong2021influenza} etc.
GNN in IoT sensing has been met with remarkable success with exceeding or comparable performance to existing benchmarks. Knowing their importance, it is crucial to obtain a comprehensive understanding in new developments in GNN techniques and their applications in IoT sensing.


In this survey, we strive to make up for this gap by summarizing existing GNN works in IoT sensing applications. We create a unified framework for IoT sensing paradigms, which divides IoT sensing into human sensing, autonomous things, and environmental sensing. We then provide a brief introduction to GNN, and define a taxonomy for GNN modeling in IoT sensing. The GNN methods are then categorized into multi-agent interaction, human state dynamics, and IoT sensor interconnection according to how the data are represented by graphs to model the problems. Through this taxonomy, we provide a comprehensive review of the most important works of GNN models in IoT sensing. In addition, we pull together a list of public data resources that are accessible for research in the intersection of GNN and IoT sensing, as well as a summary of different IoT application areas using GNN-based techniques. Lastly, we provide an in-depth comprehensive discussion on important challenges in this area and point out the corresponding future directions for research.

The remainder of this article is organized as follows: Section \ref{previous_work} illustrates how our survey is different from the existing ones in GNN methods. Section \ref{paradigms_gnn} establishes IoT sensing paradigms to sort out different sensing objects in IoT systems and how they are related to each other. Then, it introduces the preliminary knowledge about graph neural network to help readers' understanding in key GNN concepts. Building on Section \ref{paradigms_gnn}, Section \ref{taxonomy_gnn} proposes a new taxonomy to categorize GNNs in IoT sensing. In Section \ref{data_application}, we summarize the existing public data sources that are accessible and the existing applications in IoT sensing. In Section \ref{challenge_future}, we discuss existing challenges in adopting GNN models in IoT sensing applications, and point out potential future directions in this field. We close this survey paper with a conclusion in Section \ref{conclusion}.

\section{Previous Works}\label{previous_work}

There are several research review papers studying the theories and methodologies of GNNs as well as their applications on real world problems, but few works systematically summarize the applications of GNN models in IoT sensing solutions. We present some relevant representative works, and illustrate the differences between them and our current survey focusing on GNNs in IoT.

From 2020, with the boom of GNNs and their applications, several surveys reviewed the different categorizations of GNNs and their underlying theories. Wu \emph{et al.} \cite{wu2020comprehensive} proposed a new taxonomy to divide the recent GNNs into four categories, namely recurrent graph neural networks, convolutional graph neural networks, graph autoencoders, and spatial-temporal graph neural networks. Similarly, Wu \emph{et al.} \cite{wu2020graph} categorized existing GNNs into five paradigms, which include graph recurrent neural networks, graph convolutional networks, graph autoencoders, graph reinforcement learning, and graph adversarial methods, based on their model architectures and training strategies. Also, Zhou \emph{et al.} \cite{zhou2020graph} presented a general design pipeline for GNNs and discussed the derived variants with respect to their propagation, sampling, and pooling processes. Waikhom \emph{et al.} \cite{waikhom2021graph} summarized GNNs in each learning setting, including supervised, unsupervised, semi-supervised, and self-supervised learning. Zhang \emph{et al.} \cite{zhang2021evaluating} conducted a systematic evaluation on the state-of-the-art GNNs to investigate the factors that cause compromised performance in deep GNNs, the application scenarios that suite the best for deep GNNS, and how we build them, with respect to accuracy, flexibility, scalability and efficiency of GNNs. Similar survey works on GNN categories and theories also include \cite{gupta2021graph,xia2021graph}. For technical issues in designing and implementing specific GNNs, various works surveyed them from the perspectives of explainability \cite{yuan2020explainability}, dynamics \cite{skarding2021foundations}, and expressive power \cite{sato2020survey}.

From the perspective of application, GNNs are widely applied to various real-world problems, ranging from traffic forecasting \cite{jiang2021graph}, social recommendation \cite{fan2019graph}, action recognition \cite{ahmad2021graph}, to natural language processing (NLP) \cite{wu2021graph}, \emph{but no survey work has yet focused on the applications of GNN in IoT}. To be specific, Wu \emph{et al.} \cite{wu2020graph} provided a taxonomy of GNN-based recommendation models according to the types of information and recommendation tasks. Jiang \emph{et al.} \cite{jiang2021graph} studied how GNNs inherently capture structural information stored in knowledge graphs to understand the strengths and weaknesses of existing GNN paradigms. Nazir \emph{et al.} \cite{nazir2021survey} studied the GNNs applied in image classification tasks, where images can be converted into superpixels that form region adjacency graphs. Wu \emph{et al.} \cite{wu2021graph} systematically organized existing research of GNNs for NLP into a new paradigm consisting of graph construction, graph representation learning, and graph-based encoder-decoder, especially for NLP problems that can be best represented with graph structures. In addition, Ahmad \emph{et al.} \cite{ahmad2021graph} summarized the applications of GNNs in human action recognition work, where graph models are applied to represent the non-Euclidean body skeleton. 

In this work, we aim to fill this gap by connecting GNNs and IoT sensing technology and discuss in depth how GNNs are leveraged to model problems using  networked IoT solutions.

\section{IoT Sensing and Graph Neural Networks}  \label{paradigms_gnn}

\begin{figure}[!t]
\includegraphics[width=0.95\columnwidth]{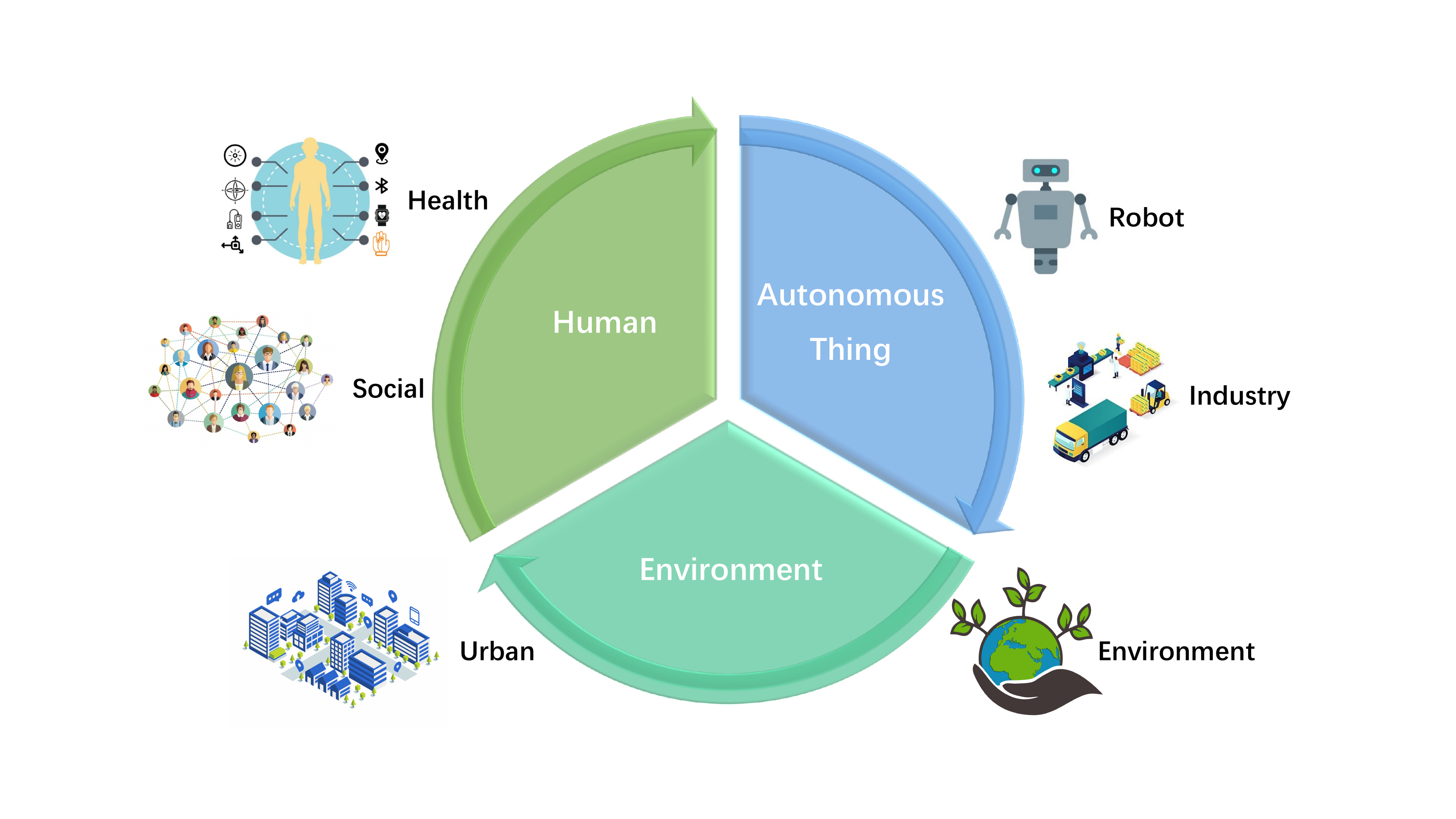}
\caption{Typical IoT sensing objects from the perspectives of autonomous thing, human, and environment.}
\label{fig:sensing-object}
\end{figure}

\subsection{IoT Sensing Objects and Paradigms} \label{iot_sensing}
Advanced IoT sensing techniques provide powerful schemes to collect rich and persistent information that characterizes real-world objects from \emph{autonomous thing}, \emph{human}, to \emph{environment}, as well as the interactions between them \cite{guo2013opportunistic}, as shown in Figure \ref{fig:sensing-object}. Given that IoT sensors, data collection schemes, data analysis methods, as well as the prospective outcomes of interest in different IoT solutions might be completely different, we review and summarize typical IoT sensing paradigms and related techniques for the different IoT sensing objects in autonomous thing, human-centric, and environment embedded systems, and their combinations.

\subsubsection{\textbf{Autonomous Things}}

IoT sensors embedded in multiple autonomous things (e.g., autonomous cars, plants, robotics, and industrial machines) provide both internal (e.g., vibration inside the machine) and external (e.g., proximity between the robots) information to further characterize the current state of operations of the targeted objects. 

IoT for \emph{robotics} and \emph{industry} refers to monitoring and evaluating both the internal operation status of individual robotics and the external interactions among group(s) of robotics and their environments through networked IoT sensors \cite{hebert2000active}. Looking at autonomous driving from the perspective of a single car, multiple sensors like cameras, Lidar, and Radar sensors, may be embedded to perceive its surroundings \cite{yeong2021sensor}. A rising research direction named internet of vehicles considers the cars on the road as connected autonomous vehicle networks, by collectively aggregating and communicating fine-grained interactions among the vehicles \cite{lee2016internet}. By centrally gathering the information of the vehicles and comprehensively managing the traffic on a vehicle network level, there is promise to improve the global efficiency of traffic. Also, IoT sensing is the necessary condition to achieve effective collaborative and cooperative work of robotics in industrial production process \cite{petersen2019review}.

\subsubsection{\textbf{Human}}

Human sensing, or human-centric sensing, aims to collect and capture people's physiology, behavior, and/or mobility information unobtrusively by leveraging ubiquitous IoT sensing devices within a human environment, including ambient senors (e.g., cameras, acoustic sensors, and WiFi) and mobile devices (e.g., smartphones and wearables). Common IoT paradigms in the scope of human-centric sensing vary from mobile sensing, to ambient sensing, to social sensing, to proivde outcomes related to health and social well-being. 

For \emph{health}, with IoT sensing schemes (e.g., camera, depth, thermal, radio, and acoustic sensors), researchers have deployed a wide variety of IoT systems for healthcare objectives in both clinical settings and daily living spaces \cite{haque2020illuminating}. For instance, camera and multiple sensors-based methods applied in intensive care units (ICU) help with continuously monitoring patients' physiological indicators and activities, which relieves the burden of wearing complex medical sensors \cite{ma2017measuring,luo2018computer}. 
There are also several novel works on using commodity networked devices (e.g., WiFi and LoRa) to monitor health indicators in an off-the-shelf and non-invasive manner. Typically, WiFi devices have been adapted for at-home healthcare targets such as respiration sensing \cite{zeng2020multisense} and sleep sensing \cite{yu2021wifi}, by analyzing their channel state information (CSI). In addition, mobile sensing paradigm, which collects data from mobile devices \cite{wang2021personalized}, is being used to understand and assess one's social anxiety \cite{boukhechba2017monitoring,boukhechba2018predicting}, infectious disease risk \cite{dong2021influenza}, and disease progression \cite{sudre2021attributes}, by passively sensing people's physiology, mobility, and activity. 

\emph{Social} is a setting where humans live, communicate, and interact in a group \cite{morrison2006marx}. IoT schemes provide insightful tools to collect real-time and scalable information from dynamic cyber-physical worlds for various social settings such as social circles with different socioeconomic statuses \cite{liu2015social,pandharipande2021social}. 
Typically, IoT techniques could be leveraged to collect rich information describing people's activities, dynamics, and interactions in society. For instance, location-based services and social networks (LBSNs), are enabled by smart mobiles and wearables, which improve personalized social recommendations \cite{cho2011friendship}, location recommendations \cite{ye2010location,yang2014modeling}, and group-oriented advertisements \cite{zhang2021group}. Yang \emph{et al.} \cite{yang2019revisiting} used  a large-scale LBSN dataset to propose a hypergraph model that studies user mobility and social relationships to reveal their asymmetric impacts on each other.

\subsubsection{\textbf{Environment}}

IoT for Environment, ranging from urban environment to natural environment, is a traditional application scenario of sensors, such as traffic flow monitoring in the city and meteorological monitoring in the wild. Recently, the proliferation of IoT devices has led to new generations of technical and computational methods in various environmental scenarios \cite{campbell2006people,zhang20144w1h}.

\emph{Urban} sensing, where IoT sensors serve as the fundamental tools, empowers modern smart city and urban computing applications \cite{zheng2014urban} for urban settings such as monitoring traffic \cite{sivanathan2017characterizing} and medical \cite{wang2020demand} systems. Specifically, flexible IoT edge devices, such as mobiles and cameras, are major sources of real-time spatio-temporal data streams which describe the complex dynamics of parameters in the urban environment (e.g., traffic speed/flows, hospital visit demands, and business hot-spots). Bike-sharing systems are another typical application empowered by IoT location-based devices. By collecting the spatio-temporal distribution of bikes in the city with network GPS edge sensors in real-time, bike sharing terminal systems deployed in the cloud server could timely schedule and rebalance the bikes to achieve optimal business revenues and meet user demands \cite{liu2016rebalancing}. Also, mobile crowdsensing \cite{guo2015mobile}, as compared with traditionally deployed fixed sensors (e.g., roadside speedometer) in urban environments, provides flexibility for other sensor-enhanced mobile devices to join the sensing network voluntarily and share urban information that can be further aggregated in the cloud for urban sensing tasks without incurring costs of deployment and maintenance of additional sensors.

Sensing for \emph{environmental} factors (e.g., temperature, air quality, and soil moisture) has incorporated various specialized sensors into its solutions. The benefits of IoT techniques on environmental sensing are two-fold. First, various novel IoT sensors (e.g., Radar, Lidar, and LoRa) have been creatively adapted for the use of environmental monitoring. For instance, in smart agriculture scenarios \cite{mekala2017survey}, sensors such as LoRa and Lidar have been applied to soil moisture monitoring \cite{heble2018low}, water management \cite{khoa2019smart}, and crop monitoring \cite{liqiang2011crop}. Using LoRa LPWAN technology, Heble \emph{et al.} \cite{heble2018low} proposed a novel low-cost watering management system by replacing traditional agricultural meters with networked LoRa sensors. Second, networked IoT sensors (i.e., IoT network) and related computing techniques (e.g., cloud computing, edge computing) have led to faster communication (information is transmitted via network) and deeper monitoring (flexible IoT sensors might be deployed to a broader range), which is greatly helpful for comprehensively monitoring and understanding environmental issues such as sustainability and climate change \cite{hart2015toward,salam2020internet}.

\subsubsection{\textbf{Discussion}}

Though the sensing objects (i.e., human, thing, and environment) as well as their typical objects (e.g., health and social, robotic and industrial, urban and environmental) are reviewed separately above, they are usually overlapped. For example, in a high-level view, the interactions between human, thing, and environment are naturally existing (i.e., human-environment and human-robot interactions), and in practice the IoT sensing tasks are oftentimes not separated \cite{galvani2016human}. In detail, though \emph{health} and \emph{social} sensing paradigms might overlap in common participants (e.g., specific groups of people), technologies (e.g., IoT wearables), and/or objectives (e.g., well-being), health sensing focuses on improving individual medical care and population health outcomes \cite{wang2021personalized}, whereas social sensing aims at describing and understanding social trends, dynamics, and interests \cite{galesic2021human}.

\subsection{Preliminary of Graph Neural Networks} \label{gnn_preliminary}
\begin{center}
\begin{figure}[!t]
\includegraphics[width=1\columnwidth]{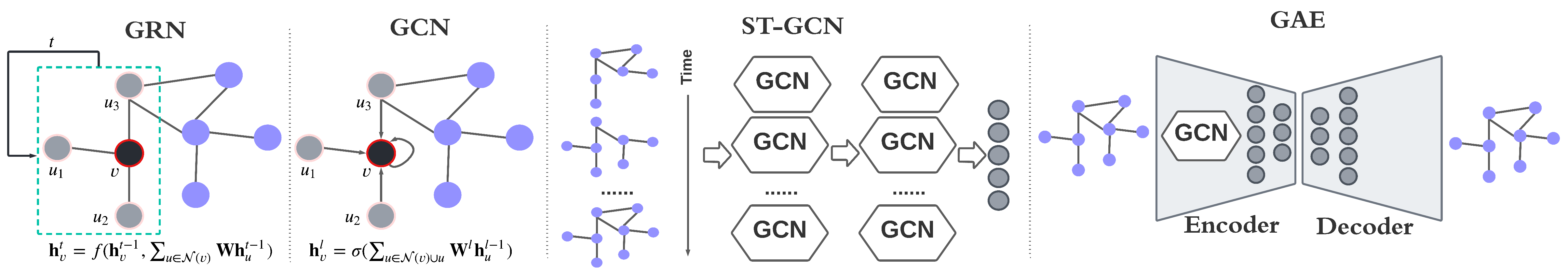}
\caption{Graph Neural Network Architectures.}
\label{fig:gnn}
\end{figure}
\end{center}

Graphs are unique data structures that consist of a set of nodes and edges that can represent numerous connected structures, such as social networks, protein-to-protein interactions, human skeletal systems, etc.~\cite{wu2020comprehensive}. Given graph structured data inputs  generated from non-Euclidean domain space, conventional DNNs (e.g., CNN and RNN) are unable to perform regular convolution operations or recurrent connection learning for long-term dependence. Graph neural networks (GNNs), on the contrary, can process graph-structured information by mapping graph inputs to numerical spaces. In this section, we provide a brief summary of current GNN models. Four different types of GNN models are shown in Fig.\ref{fig:gnn}. More comprehensive reviews of GNNs can be found in these works \cite{zhang2020deep, zhou2020graph, wu2020comprehensive}.

\subsubsection{Graph Notations and Types}
A graph is usually denoted as $\mathcal{G} = \{\mathcal{V}, \mathcal{E}\}$, where $\mathcal{V} = \{v_1,..., v_n\}$ is the set of $n = |\mathcal{V}|$ nodes and $\mathcal{E} = \{(v_i, v_j)|i \neq j, v_i, v_j \in \mathcal{V}\}$ is the set of $m = |\mathcal{E}|$ edges. Let $A \in \{0, 1\}^{n \times n}$ be the adjacency matrix where $A_{ij} = 1$ if $v_i$ is adjacent to $v_j$, otherwise $A_{ij} = 0$. Then we have the diagonal degree matrix of $\mathcal{G}$: $D$, where element $D_{ii} = \sum_{j=1}^{n}A_{ij}$ is the number of nodes that are adjacent to $v_i$. Based on the node and edge construction, graphs can be categorized as follows\cite{wu2022graph}\cite{ma2021deep}:

\begin{itemize}
    \item  Directed/Undirected Graphs: in directed graphs, edges have directional information, pointing from one node to another, which implies that messages can only transmit by following this direction. Thus, for directed graphs, $A_{ij}$ may not be equal to $A_{ji}$. In undirected graphs, there is no such direction information in the edges, implying that messages can transmit in any direction. Therefore, $A_{ij} = A_{ji}$ holds for every node in the graph and its adjacency matrix $A$ is symmetric. 
    
    \item Weighted/Unweighted Graphs: weighted graphs refer to graphs with attributed edges, which determine how messages can transit through the networks. Hence, $A_{ij} = w_{ij}$ where $w_{ij}$ is the assigned weight to edge $(v_i, v_j)$. Unweighted graphs indicate a graph with unattributed edges, in which case we can just use an adjacency matrix with 0/1 entries to describe the connectivity of graphs. 
    
    \item Homogeneous/Heterogeneous Graphs: homogeneous graphs include nodes and edges that have the same types, while heterogeneous graphs consist of nodes and edges that have different types. Therefore, heterogeneous graph can be defined as $\mathcal{G} = \{\mathcal{V}, \mathcal{E}, \phi, \psi\}$, where each node $v \in \mathcal{V}$ is associated with a node type $\phi(v)$, and each edge $(v_i, v_j) \in \mathcal{E}$ is associated with an edge type $\psi(e)$.

    \item Static/Dynamic Graphs: the topological structures of graphs cannot vary through time changes in static graphs. In dynamic graphs, their topological structures can change with time. Therefore, if a dynamic graph has $T$ graph snapshots $\{\mathcal {G}_0,...,\mathcal{G}_{T}\}$, $\mathcal{G}_t$ may be different from one another.
    
\end{itemize}

\subsubsection{Graph Signal Processing}
For IoT sensing, data are often in high dimensional non-Euclidean space. Therefore, the problems within IoT sensing can be formulated as signal processing problems on graphs. By adapting classical signal processing techniques (e.g., Fourier transform, filters, convolutions) to graph-structured data, graph signal processing can lay the background for graph neural networks with spectral methods. Leveraging the characteristics of their spectral domain, convolutions in the spatial domain of the graphs are simplified as multiplications in their spectral domain, and some other features of the graphs such as smoothness are observed:

\begin{itemize}
\item Laplacian Matrix and Graph Fourier Transform: The Laplacian matrix of a graph $\mathcal{G}$ is defined as $L = D - A$. Laplacian measures the divergence of the gradient of a scalar function. Similarly, Laplacian matrix of a graph $\mathcal{G}$ measures the variations of graph signals among nodes, and thus locally measures the smoothness of the graph $\mathcal{G}$. The normalized Laplacian matrix is $L = I_n - D^{-1/2}AD^{-1/2}$, where $I_n$ is an $n \times n$ identity matrix. $L$ is a symmetric positive semi-definite matrix, so we can perform eigen-decomposition to $L$. Then we have $L = U \Lambda U^T$, where $U = [u_1,...,u_n] \in \mathbb{R}^{n \times n}$ is a unitary matrix where each column is an eigenvector of $L$. $\Lambda = diag(\lambda_1,...,\lambda_n)$, and each non-zero element on the diagonal is the corresponding eigenvalue of an eigenvector, and these eigenvalues are frequency components. $U$ can be used to perform graph Fourier transform given a graph signal $x$: $\hat{x} = U^Tx$, and its inverse graph Fourier transform is $x = U\hat{x}$\cite{li2021graph}. 

\item Graph filter and convolution on graph: According to the definition of graph Fourier transform, filtering graph signals in the spectral domain is equivalent to keeping and attenuating frequency components in $\Lambda$. Thus, graph filter can be defined as the function of $\Lambda$: $\hat{g_{\phi}}(\Lambda)$, and the function is parameterized by $\phi$. Then the graph convolution with respect to graph filter and graph signal in its spatial domain is $g_{\phi}(L) * x = U\hat{g_{\phi}}(\Lambda)U^Tx$. By applying the graph filter to the Laplacian matrix, graph signals can be filtered based on learnable parameters in $\phi$. To avoid the costly computation for $L$'s eigendecomposition, a commonly used filter, polynomial graph filter, can be approximated by Chebyshev polynomials \cite{hammond2011wavelets}.

\end{itemize}

\subsubsection{Graph Neural Networks}

With the ability to conduct convolution on graphs, GNNs can capture complex interactive relationships between nodes and produce high-level representations of the graph inputs. The core mechanism of GNNs is to iteratively aggregate neighborhood information for each node and then integrate the aggregated information into the original node representation through neural information propagation. Differentiated by the aggregation and node update operations, GNNs can be classified into the following categories \cite{wu2020comprehensive}:

\begin{itemize}
    \item Graph Recurrent Neural Networks (GRNs): GRNs, pioneer works of GNNs, apply recurrent neural framework to learn node representations with shared function. For example, Gated GNN (GGNN) \cite{li2015gated} applies a gated recurrent unit (GRU) \cite{chung2014empirical} as a shared parameterized function to update node representation by considering previous node hidden state and the hidden state of its neighbors.

    \item Graph Convolutional Neural Networks (GCNs):
    GCNs define a neighborhood information aggregation process to update node representation, and stack multiple graph convolutional layers to generate high-level node embeddings. GCNs can be classified into spectral-based and spatial-based. Relying on graph signal processing algorithm, spectral-based GCNs perform graph convolutions by processing the input graph signals through a set of learnable filters to aggregate information \cite{zhang2020deep}. For example, ChebNet \cite{monti2017geometric}  approximates the spectral filter by the Chebyshev polynomials of the diagonal matrix of eigenvalues, rather than explicitly computing the graph Fourier transform. Spatial-based GCNs consider nodes' spatial relations to define graph convolutions. For example, NN4G \cite{micheli2009neural} accumulates nodes' neighborhood information as graph convolutions and uses residual connections and skip connections to reduce long-term forgetting over stacking of layers.

    \item Spatio-Temporal Graph Neural Networks (ST-GNNs): 
    Graphs in multiple real-world situations have both spatial and temporal variations, such as traffic network, social network, and skeleton-based human actions. In the dynamics of graphs, their topological structures change over time with varying node distributions and different edge connecting relations. ST-GNNs aim to model the spatial and temporal dependencies in dynamic graphs. A dynamic graph can be represented as an ordered list or an asynchronous set of graphs. In the design of ST-GNNs, GCNs and RNNs are usually integrated such that topological features can be extracted from the graph inputs by using GCNs and the temporal dependencies between graphs can be captured by using RNNs. For example, Yu et al. \cite{yu2017spatio} propose Spatio-Temporal Graph Convolutional Networks (STGCN) for traffic forecasting. STGCN integrates temporal gated convolution layers and spatial graph convolution layers to fuse features from both spatial and temporal domains.

    \item Graph Auto-encoders (GAEs): GAEs are highly correlated to graph representation learning, which aims to preserve high-dimensional complex graph information involving node features and link structures in a low-dimensional embedding space. The GAEs are similar to other types of autoencoders (AE). They both consist of an encoder and a decoder, where the encoder compresses information into a latent space and the decoder tries to reconstruct the original features or structural information from the latent space. Kipf et al. \cite{kipf2016variational} proposed a vanilla GAE/VGAE structure, which predicts the links between nodes in the reconstruction process. Modern GAE models utilize more useful information in the network like semantic context conditions \cite{yang2019conditional}, neighborhood information \cite{tang2022nwrgae}, and others \cite{pan2018adversarially, shi2020effective}.  

\end{itemize}
 
\begin{center}
\begin{figure}[!t]
\includegraphics[width=1\columnwidth]{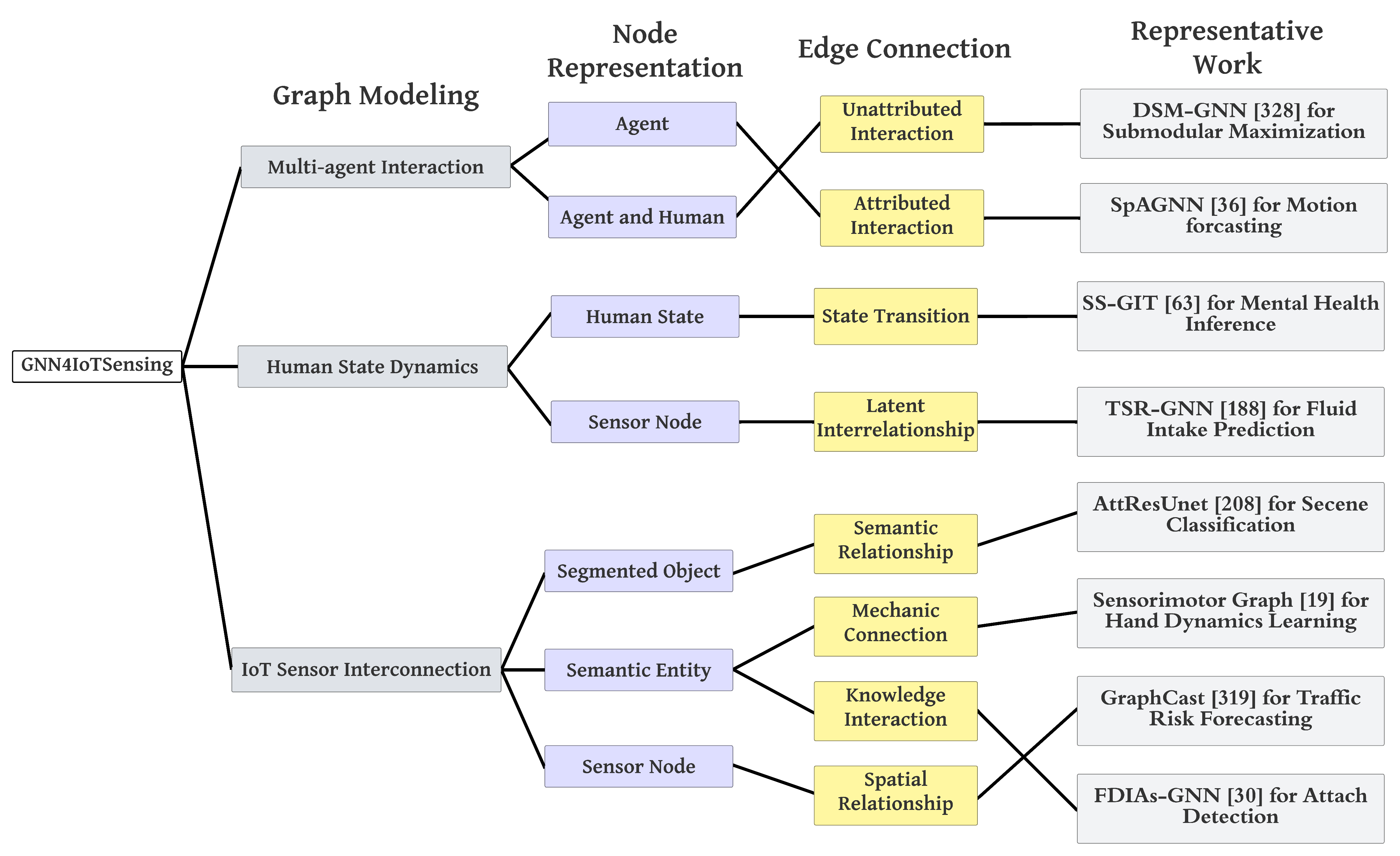}
\caption{Summary Diagram of Categorization of Graph Neural Networks in IoT.}
\label{fig:summary}
\end{figure}
\end{center}

\section{Categorization of Graph Neural Networks in IoT Sensing} \label{taxonomy_gnn}
In this survey, we categorize GNN-based models of IoT sensing into graph modeling of multi-agent interaction, human behavior dynamics, and IoT sensor interconnection. 
This GNN classification is based on the semantic explanation of graph modeling in different sensing paradigms (e.g., autonomous things, human, and environment) as mentioned in Section \ref{iot_sensing}. For example, in autonomous things, GNNs leverage graphs to represent multi-agent interactions, where nodes indicate autonomous agents or humans, and edges imply the interactive relations within thing-to-thing or thing-to-human. Our rationale behind this classification is that graph modeling of sensing objects plays a critical role in the domain application of GNNs in IoT: node representation and edge connection in graph construction have varying implications in different sensing scenarios. For example, in GNNs of multi-agent interaction, nodes represent intelligent agents, while nodes in GNNs of human behavior dynamics represent human states. The domain knowledge of the interactive relationships in IoT sensing determines the mechanism of information transitions within the sensor topology and is indicative of designing an effective and efficient GNN architecture for the corresponding learning tasks. The summary of the categorization of GNNs in IoT is shown in Fig \ref{fig:summary}.

\begin{center}
\begin{figure}[!t]
\includegraphics[width=1\columnwidth]{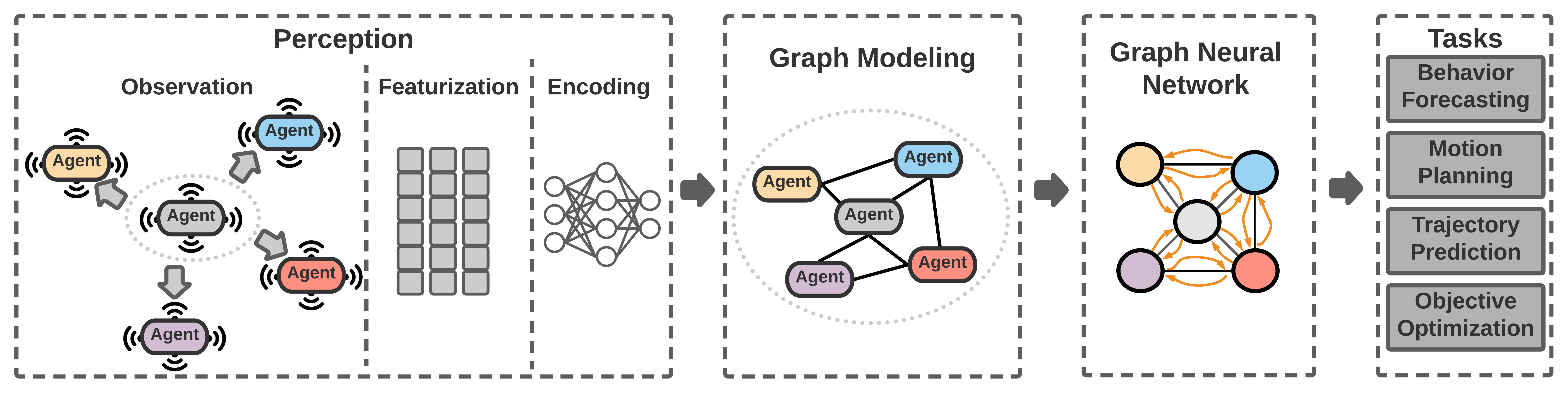}
\caption{General Graph Neural Network framework to model interactive behaviors in multi-agent systems. The nodes represent agents and the edges encode interactive relationship between agents. The framework consists of three steps: perception of environment, graph modeling of multi-agent interactions, and using GNNs to encode high level features of the neighbor agents.}
\label{fig:gnn_multiagent}
\end{figure}
\end{center}

\subsection{Multi-agent Interaction}
Multi-agent environments are ubiquitous in real-world situation, in which intelligent entities interact with each other to achieve optimized objectives. Given the interactive environment, ego-agents \cite{chiu2021encoding} (e.g., autonomous vehicles) can perceive their neighboring agents, develop knowledge of the states of other agents, and then make decisions for their individual benefits. Different from ego-agents, multi-agent systems (e.g., swarm robotics system \cite{shi2012survey}) have their inherent ability to communicate and cooperate with other agents in the systems, and solve complex problems collaboratively. By leveraging the onboard sensors (e.g., LiDAR, RGB cameras), intelligent agents are capable of detecting and recognizing their neighborhood objects, humans, and other intelligent agents, as the sensors can collect fine-grained information to profile their surrounding environment. Thus, in addition to the between-agent interactions, the relationships of intelligent agents between ambient objects and humans are also influential to their rationales of actions. Modeling object-to-agent, human-to-agent, and agent-to-agent relations and interactions plays essential roles to enable multi-agent relational reasoning in numerous research areas, such as autonomous driving, multi-agent systems, multi-object tracking, and human-robot interaction.

Extensive research has been conducted to model multi-agent interaction (MAI) in the research domains mentioned above. In autonomous driving, to navigate crowds safely, autonomous vehicles are expected to learn to predict motion trajectories of pedestrians and on-road vehicles accurately and efficiently \cite{mohamed2020social}. Traditional methods in trajectory prediction are mainly based on defining
handcrafted traffic rules and parameterized kinematic models, which are shown with limited capacity for long-term predictions \cite{jo2021vehicle}. With the advancement of deep learning techniques, current state-of-the-art methods rely on recurrent neural network that can capture high-level features from human/on-road vehicles interactive motions, and learn the parameters based on a data-driven approach. Social-LSTM \cite{alahi2016social} produces concatenated hidden states of neighboring pedestrians in a crowd and applies a social pooling layer to encode pedestrian-to-pedestrian and pedestrian-to-agent interactions for trajectory prediction. Social-GAN \cite{gupta2018social} proposes a Generative Adversarial Networks (GANs) model that consists of a RNN Encoder-Decoder generator and a RNN-based encoder discriminator to model social interactions for trajectory prediction. In Social-GAN, the generator produces trajectories with human-to-human social manners and the discriminator will learn social interaction rules and detect fake trajectories that are unconventional. Even though these RNN based models can encode multi-agent interactions implicitly into the final representation by concatenating the hidden states of each adjacent neighbors, the dynamic interactive relationships between humans and agents and inter-agent spatial topological variations cannot be utilized. In multi-agent systems (MAS), multi-agent interactions are demonstrated as collaboration and/or cooperation to accomplish shared and/or individual goals. Uniting as teams, multi-agent systems have been investigated to resolve complicated challenges, such as multi-robot coverage problems, multi-robot path planning, and multi-view visual perception problems. To tackle the problems in MAS, most existing works focus on algorithm design based on heuristics. For example, in multi-robot path planning, all robots navigate their behaviors by real-time on-board centralized controllers that can monitor robots' positions and destinations, and return plans of moving trajectories with coordination, as discussed in these works \cite{luna2011efficient, liu2012centralized, matoui2020contribution}. However, generating optimal solutions by heuristic algorithms can be computationally expensive and are hardly scalable in real world situations. Typically, inter-robot communications are not considered in the heuristic learning systems.

To mitigate the above limitations in the modeling of multi-agent interaction in rich sensor environments, GNNs provide a promising solution to encode the dynamic and complex interactions between agents by generating high level representation of the interactive relationships, treating each intelligent entity as nodes, and using edges to link every pair of adjacent entities. Combined with the graph structured relational information, detailed agents' states are fed into GNNs to generate final representations by aggregating neighborhood information. Various established works \cite{casas2020spagnn, li2021interactive, lee2019joint, zhou2021graph,tolstaya2020multi,jo2021vehicle,rangesh2021trackmpnn,ma2021continual,mo2020recog,kosaraju2019social,dong2021multi,weng2021ptp,chen2021spatial,li2021attentional,li2020social,eiffert2020probabilistic,mohamed2020social,zhou2022multi} demonstrate the superior performance of using GNNs in manifold research domains that involve multi-agent interactions. Table \ref{table:gnn_mai} summarizes the sensor infrastructures, GNN models, and learning targets in the collected works. Despite the numerous implementations of GNNs in different research questions, GNN modeling of multi-agent interactions can be generalized as the framework shown in Fig \ref{fig:gnn_multiagent}. And there are three steps in this GNN framework: 1) Perception of Environment, 2) Graph Modeling in MAI, 3) Graph Neural Networks for MAI.

\begin{table*}[t]
    \centering
    \renewcommand{\arraystretch}{1.35}
  \resizebox{1\textwidth}{!}{
    \begin{tabular}{c|c|c|c|c}
    \toprule
   Sensors & Input Data
    & Models & GNN Kernels & Tasks  \\
     \midrule
      LiDAR & Point clouds & SpAGNN\cite{casas2020spagnn} &MPNN\cite{scarselli2008graph}& Motion forecasting
\\
      LiDAR, Camera & 3D-Bound Boxes & GNN-LSTM\cite{li2021interactive} &  MPNN\cite{scarselli2008graph}&Trajectory prediction
\\

    GPS &  Waypoints & variant of GNN\cite{lee2019joint}&GN\cite{battaglia2018relational}&Trajectory prediction
\\
    
      LiDAR &Position coordinates &H-GNN\cite{jo2021vehicle} &GCN\cite{kipf2016semi}&Trajectory prediction
\\
    
    RGB camera & 2D-bound boxes &TrackMPNN\cite{rangesh2021trackmpnn}&MPNN\cite{scarselli2008graph}&Multi-object tracking \\
     RADAR
 & Waypoints  & GCN+DQN \cite{chen2021graph} & GCN\cite{kipf2016semi} & Path planning
\\
  LiDAR, Camera
 & Position coordinates  & DSM-GNN\cite{zhou2021graph} &  GCN\cite{kipf2016semi} & Submodular maximization  \\
 RADAR
 & Waypoints  & S-GNN\cite{tolstaya2020multi} & GCN\cite{kipf2016semi} & Multi-robot path planning
\\
 LiDAR
 & Position coordinates  & GNN-CVAE\cite{ma2021continual} & GAT\cite{velivckovic2017graph}&  Trajectory prediction
\\
LiDAR
 & Position coordinates  & ReCoG\cite{mo2020recog} & GAT\cite{velivckovic2017graph}&  Trajectory prediction
\\
Camera
 & Trajectories & Social-BiGAT\cite{kosaraju2019social} & GAT\cite{velivckovic2017graph}&  Trajectory prediction
\\

RGB cameras, LiDAR
 & Maps, Trajectories & DGAN\cite{dong2021multi} & GAT\cite{velivckovic2017graph}&  Trajectory prediction
\\
LiDAR
 & 3D bounding boxes & PTP\cite{weng2021ptp} & GraphConv\cite{morris2019weisfeiler}&  Multi-object Tracking
\\
LiDAR
 & Trajectories & STGNN\cite{chen2021spatial} & MPGNN\cite{gilmer2017neural}& Trajectory prediction
\\
Camera
 & Trajectories  & Social-STGCNN\cite{mohamed2020social} & GCN\cite{kipf2016semi}&  Trajectory prediction
\\
LiDAR
 & Trajectories  & GVAT\cite{eiffert2020probabilistic} & GAT\cite{velivckovic2017graph}&  Trajectory prediction
\\
LiDAR
 & Trajectories  & Social-WaGDAT\cite{li2020social} & GDAT\cite{li2020social}&  Trajectory prediction
\\
LiDAR
 & Trajectories  &  Attentional-GCNN\cite{li2021attentional} & GCN\cite{kipf2016semi}&  Trajectory prediction
\\
RGB camera
 & 2D images  &  MCP-GNN\cite{zhou2022multi} & GAT\cite{velivckovic2017graph}&  Multi-view visual perception
\\  
   \bottomrule
\end{tabular}
}
\caption{Summary of Graph Neural Networks to Model Multi-agent Interactions}
\label{table:gnn_mai}
\end{table*}

\subsubsection{\textbf{Perception of Environment}}
The surrounding environment of intelligent agents are populated with diverse entities, such as static and dynamic obstacles, other moving agents, human beings, etc. Perception of the surroundings is the fundamental task for developing intelligent systems with the capability of automatic navigation, path planning, and control \cite{fayyad2020deep}. Sensors that initially capture data representations of the environment can create both a perceptive and a locational view of the environment \cite{campbell2018sensor}. Perception systems that are fashioned out of diverse sets of sensors enable intelligent agents to acquire knowledge about their surrounding environments. The perceived environment is represented by the sensor data that are used as inputs for intelligent agents to make decisions with autonomy.

Sensors can be generally classified into two categories according to their operational principles \cite{yeong2021sensor}. The first type of sensors is proprioceptive sensors in the autonomous systems, which function by collecting the dynamical state and internal measurements of the intelligent agents/systems (e.g. velocity, acceleration, joint angles, etc). Proprioceptive sensors include Global Positioning System (GPS), gyroscopes, gyrometers, and accelerometers. The second type of sensors is exteroceptive sensors that are used for detecting, measuring and representing external information that may have an effect on the behaviors of intelligent agents. By leveraging the exteroceptive sensors, intelligent agents are capable of recognizing surrounding objects (e.g, road marks, traffic signs), human beings, other agents and the states (position, velocity, acceleration, etc.) of the other agents. Cameras (Complementary Metal-Oxide-Semiconductor (CMOS), Infrared, and cyclops), radio detection and ranging (RADAR), and light detection and ranging (LiDAR) are representative exteroceptive sensors \cite{fayyad2020deep}. In addition to this categorization, sensors forming perception systems can also be classified as either passive sensors or active sensors. Passive sensors generate outputs by monitoring the surrounding energy using cameras and GPS, while active sensors generate outputs by transmitting energy and measuring the reflected energy (LiDAR and RADAR). 

As shown in Table \ref{table:gnn_mai}, the authors of the surveyed papers use single LiDARs/carmeras, RADAR, and a combination of cameras and LiDARs to generate data representations of other intelligent agents/human beings in the perceived local environments. Cameras use passive light sensors to provide a high-resolution digital pictures of surrounding areas. By using computer vision techniques, it is possible to detect and recognize dynamic and static objects within a broad range of areas. The advantage of cameras in perception of environment is that cameras, which approximate human visual perception capacity, can provide contextual information about surroundings, such as road signs, traffic lights, moving human beings. On the flip side, cameras are computationally expensive relative to active sensors, limited in measuring distances, have degenerated performance in poor weather conditions, and are sensitive to changes in lighting conditions. LiDARs can actively illuminate the surroundings by emitting lasers, and time measurements between sending and reflecting a pulsed laser are used to generate 3D representations of the surrounding environment. One of the benefits of LiDARs is that it can be used to provide precise and accurate localization and mapping, since they can produce a high-resolution densely spaced network of elevation points, referred as point clouds. RADARs send out radio waves that detect objects and measure their distance and velocity to the RADARs in real time. Unlike cameras, weather conditions cannot impact the performance of RADARs, and they are less expensive than LiDARs. In practice, perception systems are usually constructed from multiple sensors to compensate for individual deficiencies. For example, Daraei et al. combine LiDAR and camera data with sensor fusion techniques to decrease uncertainty and enhance detection performance \cite{daraei2017velocity}. To enhance the object detection and classification pipeline, LiDAR and camera data can also be fused together for extrinsic calibration  \cite{li2021attentional}.

\subsubsection{\textbf{Graph Modeling of Multi-agent Interaction}}\label{section4}
Graph modeling plays an essential role in capturing multi-agent interactions in numerous multi-agent scenarios, such as autonomous driving and multi-agent systems. Varying from different multi-agent scenarios, node and edge representations in graph modeling with different types of interactions have different semantic explanations. For example, in autonomous driving, nodes can be used to represent human beings or other vehicles, edges can encode the dynamic distances between two agents on the roads, and graph modeling, in this situation, aims to capture the social-awareness relationship. Differently, in multi-agent systems, graph modeling usually represents the communicative and collaborative relationships within a group of agents to achieve a unified goal. In general, graph modeling for MAI consists of node representation and edge connection.

\noindent\textbf{Node representation}: in a multi-agent environment, nodes usually represent static/non-static agents, such as human beings, other moving agents, and obstacles. With the perception systems, agents evolve with the capacity to automatically detect and track other objects in their surrounding environment. Given the sensory data generated from the perception systems, deep learning based object detection and tracking models can be applied to extract the interactive agents, moving trajectories, and other attributes (e.g., velocity, direction, between-agent distance). For example, Li et al. realized 2D object detection and classification by using YOLOv3 and a 3D LiDAR classifier called PointNet respectively, and applied SORT, a Deepsort visual tracker, to generate trajectories of the classified objects \cite{li2020attentional}. More detailed information about object detection and tracking can be found in these following papers \cite{jiao2019survey, pal2021deep, arnold2019survey}. Combined with node representation of detected objects, node attributes are also featurized to describe the states of agents. The basic node features can be directly extracted from the tracked trajectories. Spatial coordinates of the objects in the agents' world coordinate frames are used as node attributes in \cite{mohamed2020social, eiffert2020probabilistic, li2020attentional}, providing positional information and relative distance between agents. Authors of the works \cite{chen2021spatial, li2020graph, tolstaya2020multi} use manually designed features to describe the state of each node at each time step, such as position, velocity and acceleration. These handcrafted features can provide more detailed information about the agents' state from both spatial and temporal perspectives. The most popular node feature extraction method is using Neural Networks (e.g., Multi-layer Perceptron (MLP), Recurrent Neural Networks) to encode descriptive node features and generate high-level embeddings of agents' states. As demonstrated in these works \cite{li2020social, weng2021ptp, kosaraju2019social, mo2020recog, ma2021continual, jo2021vehicle, lee2019joint, li2021interactive, dong2021multi, casas2020spagnn, zhou2022multi, zhou2021graph, li2021message}, researchers use MLPs and/or Long-short Term Memory (LSTM) neural networks to encode motion and location information of agents and then produce feature representations that capture spatial and temporal variations as the node attributes.
 
\noindent\textbf{Edge connection}: edge information defines how the agents interact with each other in their situated environments. Beyond node representation of agents, adding edges between every two connected/communicative agents can explicitly express multi-agent interactions by using graph structured information. The edges can indicate the influence between agents, where the movement of one agent can impact the movement of other agents. In the surveyed papers, the authors of the works \cite{li2020social, eiffert2020probabilistic, chen2021spatial, mo2020recog, jo2021vehicle, casas2020spagnn, zhou2022multi, tolstaya2020multi, li2020graph, zhou2021graph} construct edges between two agents if and only if these two agents are close to each other, since the authors assume that interactions/communications can only be effective if two agents are within a bounded distance. Adjacency matrices are used to represent the edge connections within multi-agent environments, where 1 indicates adjacency and 0 indicates nonadjacency. However, discrete adjacency matrices can be limited in representing the strength or extent of influences between two agents. To address this limit, the authors in \cite{mohamed2020social} added handcrafted features to describe the edges in the graphs to model how strongly two agents influence each other by using a kernel function. And as described in these works \cite{weng2021ptp, ma2021continual, lee2019joint, li2021interactive}, instead of using manually designed features, RNNs are used to produce latent representation of edges, which can encode temporal and local-graph structural variations of the nodes in the graphs. One of the limitations of edge embeddings generated from RNNs is that agents are treated equally when they interact with each other. However, closer agents can have stronger impacts than more distant agents. Thus, to incorporate the varying impacts from agents, attention mechanism has been investigated to generate attributed edges to represent the weighted adjacency of the interaction graphs, as demonstrated in these papers \cite{li2021attentional, kosaraju2019social, dong2021multi, li2021message, ryu2020multi}. For example,  Li et al. \cite{li2021message} applied attention mechanism to assign higher attention weights for the edges when two agents are closer with each other, under the assumption that the neighboring agents' information/messages have relatively higher importance than other faraway agents. 
 
\subsubsection{\textbf{Graph Neural Networks for Multi-agent Interaction}}
Graph representations of multi-agent interaction produce graph structured data that cannot be processed by using traditional deep learning models, such as MLPs and CNNs. and GNNs. Particularly, the kernels of GNNs play an essential role in encoding the graph structured data and generating comprehensive embeddings that capture the complex interactions between agents by aggregating neighborhood information. As illustrated in Table \ref{table:gnn_mai}, in general, there are three types of GNN kernels that have been adopted in modeling MAI: 1) by relying on message passing mechanisms, the authors of the works \cite{casas2020spagnn, li2021interactive, lee2019joint, rangesh2021trackmpnn, chen2021spatial} apply MPNN/GN to model the interactions between agents; 2) based on graph convolution algorithms, the authors of the works \cite{jo2021vehicle, chen2021graph, zhou2021graph, tolstaya2020multi, weng2021ptp, mohamed2020social, li2021attentional} employ graph convolutions to generate high level representations of multi-agent interactions; 3)  by leveraging attention mechanisms, the authors of the works \cite{zhou2022multi, li2020social, eiffert2020probabilistic, dong2021multi, kosaraju2019social, mo2020recog, ma2021continual} make the neural information propagation among neighborhood with selective focus. 

\noindent\textbf{MPNN/GN-based GNNs}: in this type of GNNs, given an input of graph structured data of MAI and agent states, MPNNs/GNs activate a message passing algorithm with finite-steps traveling through the graphs to update agent states. During the message passing stage, agent state information passes through the edges to their neighboring agents and is aggregated by using a permutation-invariant function, such as summation. The aggregated representation is then integrated with existing agent state information via neural networks (e.g., MLP), resulting in updated agent states. For example, Li et al. \cite{lee2019joint} encoded previous agent states, agent-wise features (current agent states) and adjacency matrix, and used a stack of two vanilla  GN layers to produce interaction-aware agent representations for trajectory prediction. Casas et al. \cite{casas2020spagnn} included spatial feature maps, relative local coordinates, and features extracted from original coordinate system as agent states, and used the message passing algorithm to update the agent state. For the message passing process through any directed edge $(u,v)$ between agent $u$ and $v$, at propagation step $k$, the message $m^{(k)}_{u\rightarrow v}$ can be computed by 
\begin{equation}
    m^{(k)}_{u\rightarrow v} = \mathcal{E}^{(k)}(h_{u}^{k-1},h_{v}^{k-1},\mathcal{T}_{u,v}(o_u^{k-1}),o_v^{k-1}, b_u,b_v)
\end{equation}, where $\mathcal{E}^{(k)}$ is a 3-layer MLP and $\mathcal{T}_{u,v}$ is the transformation from the coordinate system of detected box $b_u$ to the one of $b_v$. $o_u$ and $o_v$ denote the relative local coordinates for the agents. Even though MPNN/GN-based GNNs can produce competitive results in trajectory prediction and motion forecasting, they can suffer from scalability issue, since processing messages of all involved agents are required by them. 

\noindent\textbf{GCN-based GNNs}: instead of aggregating node signals from spatial domains in MPNNs/GNs, GCN-based GNNs learn MAI representations from spectral domains. GCNs use graph convolutions to aggregate neighborhood information and then update each node representation, as mentioned in Section \ref{gnn_preliminary}. By considering the topology in multi-agent interactions, Zhou et al. \cite{zhou2021graph} and Weng et al. \cite{weng2021ptp} implemented graph convolutions for node aggregation to capture multi-robot communications and vehicle interactions, respectively. In addition to encoding the topological variations, some researchers also incorporate temporal variations in their proposed GNNs. For example, the authors in \cite{mohamed2020social} use a sequence of graphs to represent the social interactions varying through time changes, and feed the graphs into a spatio-temporal Graph Convolution Neural Network (ST-GCNN) to extract spatiotemporal node embeddings from the input graphs. TXP-CNN is connected after ST-GCNN to extract temporal correlations within the graph embeddings. Li et al. \cite{li2021attentional} applied the same GNN architecture to capture the spatio-temporal variations in the sequence of graphs by using attentive features for each agent in the graphs.

\noindent\textbf{GAT-based GNNs}: the core component of using attention mechanisms in GNN for MAI is based on Graph Attention Network (GAT) \cite{velickovic2017graph}, under the assumption that neighbors have varying impacts to the agents' behaviors. GATs exploit self-attention to produce node representations by putting more attention on the important features when prediction is made. During the aggregation process, more weights will be assigned to the neighborhoods that have more influences to the node. The graph attention convolution can be defined as

\begin{equation}
    h_v^{(l+1)} = \sigma (\sum_{u \in \mathcal{N}_{(v)}}\alpha_{v,u}^{(l)}W^{(l)}h_u^{(l)}),
\end{equation}
where $h_v^{(0)} = x_v$, and $\alpha_{v,u}^{(l)}$ is the attention of node $v$ in the $l$th layer, which can be expressed as
\begin{equation}
    \alpha_{v,u}^{(l)} = \frac{exp(\eta(\mathbf{a}^{(l)T}[W^{(l)}h_v^{(l)},W^{(l)}h_u^{(l)}])}{\sum_{u^{'}\in \mathcal{N}(v)}exp(\eta(\mathbf{a}^{(l)T}[W^{(l)}h_v^{(l)},W^{(l)}h_{u^{'}}^{(l)}])}.
\end{equation}

Here $\eta$ is LeakyReLU activation function, while $\mathbf{a}$ is a weight vector to parametrize the attention mechanism. Kosaraju et al. \cite{kosaraju2019social} proposed a graph-based
generative adversarial network with physical and social attention to better model the social interactions of pedestrians: multiple GAT layers are stacked to generate the agent state embeddings that contain interaction awareness in crowds. Instead of using homogeneous graphs to represent MAI, Mo et al. \cite{mo2020recog} employ heterogeneous graphs to include both agent information and context information and then apply GAT to model MAI for vehicle trajectory prediction. Li et al. \cite{li2020social} proposed graph double-attention network that can put both topological and temporal attention in node aggregation to produce high level node embeddings. 

\begin{center}
\begin{figure}[!t]
\includegraphics[width=1\columnwidth]{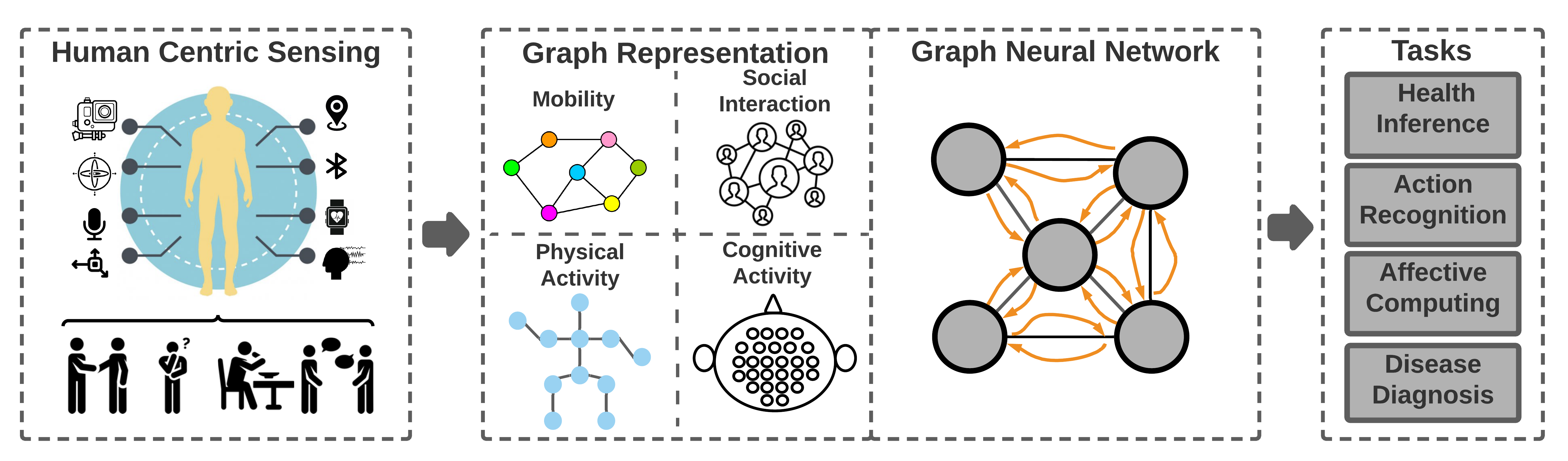}
\caption{General Graph Neural Network architecture to model human behavior dynamics by human centric sensing techniques.}
\label{fig:gnn_human}
\end{figure}
\end{center}

\subsection{Human State Dynamics}
There is currently a substantial amount of evidence demonstrating strong correlations between human states and internal (e.g., physiological signals) and/or external (e.g., ambient environment) stimuli. Human behaviors are complicated, personal, interconnected, change over time, and are impacted by each individual's historical human states.  Our behaviors occur while we go about our daily lives, and are both a reflection of our responses to and responses from the external environment \cite{sabharwal2017bio}. Human state dynamics (HSD) indicate the fluctuation of physiological and psychological human states, which enable researchers to extract bio-behavioral signatures, and provide the basis for human related inference. With the advancement of sensing techniques (e.g., smart ubiquitous devices with embedded sensors), there is increasing interest in the research of human behaviors by leveraging ubiquitous sensors in both industry and academia. Human states can be measured and monitored continuously and unobtrusively using ubiquitous devices. In the analysis of linking human states with health and well-being, fine-grained behavioral indicators can be engineered and generated from the sensory data to provide valuable information of vital biometric responses and behavioral patterns in natural settings \cite{hammal2020face}. In this vein, Dong et al. proposed to use mobile sensing to record people's behavioral footprints during the COVID-19 pandemic and get a better understanding of their collective behavior changes in response to COVID-19 regulations \cite{dong2021detection}. By using mobile sensing techniques to passively collect behavior related information, students’ mobility, activity levels, and communication patterns can be associated with their social anxiety, depression and affect levels \cite{boukhechba2018demonicsalmon}. As a noninvasive alternative of neuroimaging scans (e.g., MRI and PET), electroencephalography (EEG) shows its potential for Alzheimer’s disease diagnosis \cite{cassani2018systematic}. These breakthroughs in ubiquitous sensing have led to  wide adoption of machine learning algorithms for human-centric applications.

Unlike natural language processing and computer vision, the major challenge of modeling human-centric sensor data is its multi-modality and multi-channel sources, requiring properly constructed deep learning algorithms to efficiently fuse these heterogeneous data sources for superior performance.
According to Liu et al., a novel multiplicative multimodal method has been suggested to select more significant data modalities on a sample basis~\cite{liu2018learn}. A unique approach for autonomously selecting mixtures of modalities was also assessed as an extension of their findings, to capture any cross-modality correlations and complementarity.
Chakraborty et al. developed a multi-channel convolutional neural network architecture for identifying states of mind utilizing a variety of physiological data sources, including ECG, accelerometer, and respiration data from ubiquitous devices~\cite{chakraborty2019multichannel}.
Kampman et al. used a trimodal architecture to blend video clips, audio, and text data to predict Big Five Personality Trait scores~\cite{kampman2018investigating}. They used stacked convolutional neural layers on each data modality and concatenated the outputs of all channels with fully connected layers to produce a 9.4\% improvement over the best individual modality model.
Li et al. preprocessed multichannel EEG data into grid-like frames and created a hybrid deep learning model using CNN and RNN layers~\cite{Li2016emotion}. The researchers used this method in emotion identification tasks for the purpose of diagnosing emotional disorders in neurology and psychiatry. A similar CNN+RNN architecture was applied by Salekin et al. in the evaluation of newborn pain using video data, combining facial expressions and body movement signals into a three-channel model consisting of face, body, and face+body hybrid channels, as well as other applications~\cite{salekin2019multi}. Despite the superior performance of using CNN and/or RNN in modeling the multi-modal data, complex interconnection and interdependence between human behaviors and sensors cannot be appropriately represented by using the conventional deep learning methods.

GNNs have demonstrated superior performance while overcoming the limitations of conventional deep learning methods when modeling the heterogeneous sensor data. The core motivation of using GNNs for modeling human behavior dynamics is that graph structured information can capture complex interactions among human behaviors by generating explicit topological representations to enhance the expressive power of the sensory data to further improve the prediction performance. Various works have demonstrated improved performance by applying GNNs in modeling human behaviors using the sensory data  compared to other non-GNN approaches, particularly in domains such as social interaction detection, mobility prediction, cognition and physical activity recognition  \cite{dong2021influenza} \cite{dong2021using} \cite{jalata2021movement}  \cite{ngoc2020facial} \cite{shi2019skeleton} \cite{feng2021multi} \cite{li2021cross} \cite{yu2021accurate} \cite{guo2020sparse} \cite{zhong2020eeg} \cite{zhao2020spidernet} \cite{jalata2021movement} \cite{li2021multiscale} \cite{zhang2019graph} \cite{zhao2019bayesian} \cite{parsa2020spatio} \cite{li2020dynamic} \cite{yang2021shallow} \cite{song2018eeg} \cite{li2019classify} \cite{li2021cross} \cite{lun2020gcns} \cite{jang2018eeg} \cite{wagh2020eeg} \cite{kwak2020graph} \cite{lihuman2021} \cite{zhao2020spidernet} \cite{liu2020handling} \cite{si2019attention}  \cite{li2019classify} \cite{dong2021semi}. We generalize the process of using GNNs to model HSD as a framework and show it in Fig. \ref{fig:gnn_human}. GNN modeling for HSD consists of three steps: 1) human centric sensing; 2) graph representation of human behaviors (sensors); 3) behavior inference. Table \ref{table:gnn_hbd} summarizes the differences between existing works, including the sensors used, the input data type, the GNN models, the GNN kernels and learning targets.
 
\subsubsection{\textbf{Human-centric Sensor Network}}

Human-centric sensing systems involve a greater degree of human participation at additional stages along the data-to-decision process. This path typically entails sensing (i.e. acquiring sensor measurements) and information processing, which consists of extracting useful information from sensor measurements, analyzing it, and deriving knowledge from it \cite{srivastava2012human}. Advances in ubiquitous sensing have enabled us to observe human subjects with minimal interference, making it possible to investigate the patterns of human behavior dynamics with respect to internal and/or external influences. Human-centric sensing systems can be categorized into body sensor network (BSN), rich-sensor mobile computing devices (MCD), and a combination of BSN and MCD. BSN is a collection of low-power and lightweight wireless sensor nodes for monitoring human body processes and the surroundings. BSNs have been used to track users' activity in a number of situations, including but not limited to illness identification and prevention \cite{crispim2017online} via activity analysis, rehabilitation after a medical operation \cite{yu2018non}, and emotion detection of drivers \cite{rebolledo2014developing}. For example, Ali et al. demonstrate the success of using wearable bio-sensors (e.g., Electroencephalogram (EEG)) to develop a robust emotion recognition model for patients with special needs in ambient assisted living environments\cite{ali2016eeg}. MCD is a ubiquitous system that is formed by a diverse set of embedded sensors (e.g., accelerometer, gyroscope, GPS), that continuously and unobtrusively collects human biobehavioral footprints. Numerous mobile sensing systems have been employed for anxiety recognition \cite{boukhechba2017monitoring}, symptom detection \cite{dong2021influenza}, and fluid intake management \cite{tang2022sensor}. For example, by leveraging the embedded sensors in smartphones and smartwatches, raw mobile sensing data can be translated into behavioral markers and linked to mental health states \cite{wang2021personalized}.

Ubiquitous devices can continuously and passively record human state variations through high quality measurements. Their embedded sensors have different sensing capabilities, transmission standards, storage, and computation power demands \cite{fortino2012enabling}. Numerous aspects of human states have been extensively investigated and proven to provide valuable connections between sensory data and human related results (e.g., health state, emotions, and biometrics). There are two types of signals generated from human-centric sensing: behavioral signals and biomedical signals. Behavioral signals explain human behavioral changes in response to various types of stimuli, which include social interaction signal, motion signal, mobility signal, and communication signal. For example, in motion sensing, human body movements can be tracked by sensors such as accelerometers, pedometers, gyroscopes, and magnetometers \cite{jebelli2016fall}. Bluetooth and RFID technologies, as well as use the of a microphone have been implemented in a smart environment to detect and measure the level of social interactions among people \cite{masciadri2019detecting}. An additional example is the use of a smartphone's embedded GPS to extract mobility features as a digital biomarker to correlate with negative symptoms in schizophrenia \cite{depp2019gps}. Biomedical signals include physiological and biomechanical variables, such as electrocardiogram, oxygen saturation, blood pressure, body temperature, respiratory rate, heart rate or blood glucose concentration, skin conductivity, and electrodermal activity. 
For example, in addition to accelerometer, Empatica E4 wristband contains an electrodermal
activity (EDA) sensor, and a photoplethysmography (PPG) sensor to measure blood volume pulse (BVP) \cite{seneviratne2017survey}. Electrocardiogram (ECG) devices may be used to monitor heart rate: the electrical activity of a patient's heart is measured through electrodes placed on various locations of his/her body \cite{phan2015smartwatch}. By measuring the voltage difference between two electrodes placed on the surface of human scalps, EEG sensors can trace the electrical activity of the brain \cite{gu2021eeg}.

\begin{table*}[t]
    \centering
    \renewcommand{\arraystretch}{1.35}
  \resizebox{1\textwidth}{!}{
    \begin{tabular}{c|c|c|c|c}
    \toprule
   Sensors & Input Data Type
    & Models & GNN Kernels & Tasks  \\
     \midrule
      Acce,GPS,Gyro,Blth & Multi-modal & SS-GIT\cite{dong2021semi} &GIT\cite{dong2021semi}& Mental Health Inference
\\
      EEG & Uni-modal & ST-GCNN\cite{li2019classify} &  GCN\cite{kipf2016semi}&Neuroexcitation Classification
      \\
GPS,Gyro,PPG,Light
 & Multi-modal & SRDA\cite{tang2022sensor} & GCN\cite{kipf2016semi}&  Anomaly Detection
\\

    EEG &  Uni-modal & GCNN\cite{jang2018eeg}&ChebNet\cite{defferrard2016convolutional}&Video Identification
\\

Blth,Acce,Mag,Gyro 
 & Multi-modal & GraphConvLSTM\cite{han2019graphconvlstm} & GFT\cite{shuman2013emerging}&  Activity Recognition
 \\
      EEG &Uni-modal &DGCNN\cite{song2018eeg} &GFT\cite{shuman2013emerging}&Emotion Recognition
\\
    
    EEG & Uni-modal &GCNN\cite{wagh2020eeg}&GCN\cite{kipf2016semi}&Neurological Disorder Diagnosis \\
    
     Acce,Incl,Light
 & Multi-modal  & G-GRL\cite{dong2021using} & GeoScattering \cite{velivckovic2017graph}&  Cortisol Level Prediction
\\
     EEG
 & Uni-modal  & GCNs-Net \cite{lun2020gcns} & GFT\cite{shuman2013emerging} & Motor Imagery Classification
\\
  EEG
 & Uni-modal   & SOGNN\cite{li2021cross} & GFT\cite{shuman2013emerging} & Emotion Recognition  \\
 Acce,GPS,Gyro,Blth 
 & Multi-modal & Multi-GNN\cite{dong2021influenza} & GraphSage\cite{hamilton2017inductive}&  Symptom Recognition
 
\\
 EEG
 & Uni-modal  & RGNN\cite{zhong2020eeg} & GCN\cite{kipf2016semi} & Emotion Recognition
\\

GPS,Gyro,PPG,Comp,Light
 & Multi-modal & TSR-GNN\cite{tang2022using} & GIN\cite{xu2018powerful}&  Fluid Intake Prediction

\\
Acce,GPS,Gyro,Blth 
 & Multi-modal & ST-GCN\cite{dong2022incremental} & GCN\cite{kipf2016semi}&  Symptom Recognition

\\
   \bottomrule
\end{tabular}
}
\caption{Summary of Graph Neural Networks to Model Human Behavior Dynamics}
\label{table:gnn_hbd}
\end{table*}

\subsubsection{\textbf{Graph Modeling of Human Behavior Dynamics}}
The dynamics of human states contain rich information related to the transitions between two humans states: previous human states can imply the next human state while the current human state is helpful to gain knowledge about historical human states. Complex state transitions within human systems make it challenging to capture inter-state relations and translate the collected sensory data into explainable knowledge. Graph modeling can transform unstructured human state information into graph structured representation that connects individual human states effectively. In general, there are two different classes of graph modeling in HSD: 1) using graphs to generate explicit representation of human behavior (body) dynamics or latent human state transitions; 2) using graphs to represent sensor topology in which individual sensors capture a certain channel of human behaviors. 

\noindent\textbf{Node representation}: in the graph model of HSD, node representation determines how human states can be viewed as vertices in graphs. Different HSD graph modeling methods imply different semantics of the vertices in node representation. For graph modeling that uses graphs to represent latent human state transitions, each node indicates an individual hidden human state. Given multi-modal sensory data, in most of the cases, human states are hidden from the collected sensor data. For example, given continuous accelerometer data about human body motions, we can extract hidden human states from the accelerometer data where the states can represent different human body movements (e.g., running, walking). In \cite{dong2021using}, Dong et al. used clustering to generate a sequence of human state transitions from multi-channel mobile sensing data comprised of accelerometer, inclinometer, and a light sensor. Then, they denoted each cluster as a node in the behavior state transitions. Similarly, the authors in these studies \cite{dong2021influenza, dong2021semi, dong2022incremental} produced graph representations from multi-channel human behaviors, including social interaction, mobility, and physical activity, and used Bluetooth encounters as nodes in social interaction graphs, visited places as nodes in mobility trajectory graphs, and body movement as nodes in physical activity transition graphs. In the studies of monitoring brain electrical activities, electrical signals produced from segmented regions on human scalp indicate the neuron activities in the corresponding brain region \cite{michel2019eeg}. However, the performance of graph modeling based on explicit graph representations of human states (behaviors) can be impacted by the performance of the employed clustering algorithm, which increases model tuning difficulty as more hyperparameters will be introduced by using clustering techniques. To tackle this problem, Tang et al. \cite{tang2022sensor,tang2022using} represent each sensor as a node in the graph representation of multi-modal sensor data. Instead of aggregating similar sensor data points into vertices in graphs, the authors in these works included granular sensor data in the node attributes in their sensor network graph models. In the graph modeling of neuron activities based on EEG, the nodes represent segmented scalp regions and each EEG channel corresponds to a node in the graphs, as demonstrated in these works \cite{zhong2020eeg, li2021cross, lun2020gcns, wagh2020eeg, song2018eeg, jang2018eeg, li2019classify}.

\noindent\textbf{Edge connection}: edges link neighboring nodes and constructively contribute to information transmission in graphs. In different HSD graph modeling methods, edge connections have different semantic explanations. In graph representation of human behavior (body) dynamics or latent human state transitions, edge connections describe the mechanism about human state (behavior) changes in their daily lives. For example, in human mobility trajectories, 
edge connections between two nodes in the trajectories imply people traveling from one place to another, as demonstrated in the works~\cite{dong2021influenza, dong2021semi, dong2021using, dong2022incremental}. Dong et al. linked sequential human states (mobility behavior, physical activity, social interaction) to construct graphs. Human state transition graphs describe individuals' behavioral reactions to internal and/or external stimuli. 
In graph representation of sensor topology, edges connect sensor vertices in graphs to indicate the interdependence between sensors. They imply the interactions between the corresponding sensed behaviors. Different from pre-defined edge connection mechanism in the graph construction of human state transitions, there is no prior knowledge about the neighborhood relations in sensor topology. 
Numerous researches have been conducted to learn the relationships between sensors by using data driven approaches. Tang et al. \cite{tang2022using} calculated inter-correlations between sensors and selected the top K most interdependent neighbors for each sensor as adjacency. By leveraging attention mechanisms, the authors generated attention weight matrices based on multi-modal sensory data, and employed the sensor attention weight matrices to describe connections between sensors \cite{tang2022sensor}. In the research of monitoring human neuron activities using EEG, the edge connections between electrodes define the strength of connections between brain regions. Inspired by the theories of human brain organization, Zhong et al.\cite{zhong2020eeg} proposed a method to calculate inter-channel relationships by an inverse square function of physical distance between electrodes. In addition to spatial connection, Wagh and Varatharajah \cite{wagh2020eeg} measured the channel connectivity by combining spatial and functional connectivity between electrodes. 

\subsubsection{\textbf{Graph Neural Networks for Human State Dynamics}}

Human state transitions and sensor interactions can be comprehensively represented based on HSD graph modeling. Edge connections in the graph representation of HSD enable human state information to transmit through the edges. GNNs can produce comprehensive representations of HSD from the graph structure and relevant human state attributes: graph convolutions can aggregate multi-hop human state information to generate high-level embeddings that can capture complex interactions of human state transitions and connected sensors. As shown in Table \ref{table:gnn_hbd}, 
Two types of GNNs have been invented in the surveyed studies. The first type of GNNs for HSD can be characterized as static GNNs, in which the authors consider the spatial features of the topology in human states and sensor networks, as demonstrated in these works \cite{dong2021influenza,dong2021using,jang2018eeg,wagh2020eeg,song2018eeg,lun2020gcns,zhong2020eeg,li2021cross, tang2022sensor}. The second type of GNNs for HSD can be characterized as dynamic GNNs, in which the authors consider both spatial and temporal variations of the topology in human states and sensor networks, as demonstrated in these works \cite{tang2022using,dong2021semi,dong2022incremental,han2019graphconvlstm,li2019classify}.
 
\noindent\textbf{Static GNNs}: topological variations in HSD graph representations cause human state information to have different message transit pathways when graph convolutions activate neural information propagation. Graph convolutions in static GNNs encode graph signals and node information, and then map them into embedding spaces. Dong et al. \cite{dong2021influenza} applied a multi-channel graph convolution neural network to extract high level features from peoples' mobility trajectories, social interaction networks, and physical activity transitions. Jang et al. \cite{jang2018eeg} converted multi-channel EEG signals into graph signals that consist of graph structures and signal features, and used GCNN to learn the graph signals. Different from pre-calculated adjacency matrices in GCNN \cite{jang2018eeg}, Song et al \cite{song2018eeg} trained the DGCNN with a graph Fourier transform (GFT) kernel, which dynamically optimizes adjacency matrices in the learning process. To train more robust GNNs, Zhong et al. \cite{zhong2020eeg} leveraged domain adversarial training and emotion-aware distribution learning to better deal with cross-subject EEG variations
and noisy labels. In the scenarios of using GNNs for anomaly detection, Tang et al. \cite{tang2022sensor} proposed a dual graph autoencoder that can simultaneously learn latent graph structure and node feature representation from normal human states, and classify abnormal human states if the decoded embeddings are distant from normal decoded embeddings. 
 
\noindent\textbf{Dynamic GNNs}: in addition to the spatial domain of topological variations, graph structures in HSD can also vary in the time domain. After a period of continuous human centric sensing, a sequence of graphs with attributed nodes can be generated to encode topological and temporal dynamics within human states. Dynamic GNNs combine graph convolutional neural networks (GCNNs) and recurrent neural networks (RNNs) to capture the spatio-temporal characteristics of graph sequences. GCNNs aim to extract high-level topological features from the graph signal, and RNNs aim to decode the temporal interconnections in the chronologically ordered graphs. Dong et al. \cite{dong2022incremental} proposed a spatio-temporal graph neural network that processes a sequence of multi-channel graphs that represent dynamics of human behaviors over time: GCNs are used to generate human behavior graph embeddings, and then the sequence of graph embeddings is fed into long short term memory (LSTM) to decode the temporal relations within it. Han et al. \cite{han2019graphconvlstm} applied a sequence of graph convolutions and temporal convolutions with residual connection to enhance the capacity of learning long-term temporal dependency in dynamic GNNs. Without knowing the labels of each graph in a graph array, Dong et al. \cite{dong2021semi} proposed Graph Instance Transformer (GIT) to produce a bag of graph embeddings with more attention weighted graph instances and sub-graph structures. GIT enables training dynamic GNNs given sequences of graph structured information under weak supervision. 
 
\begin{center}
\begin{figure}[!t]
\includegraphics[width=1\columnwidth]{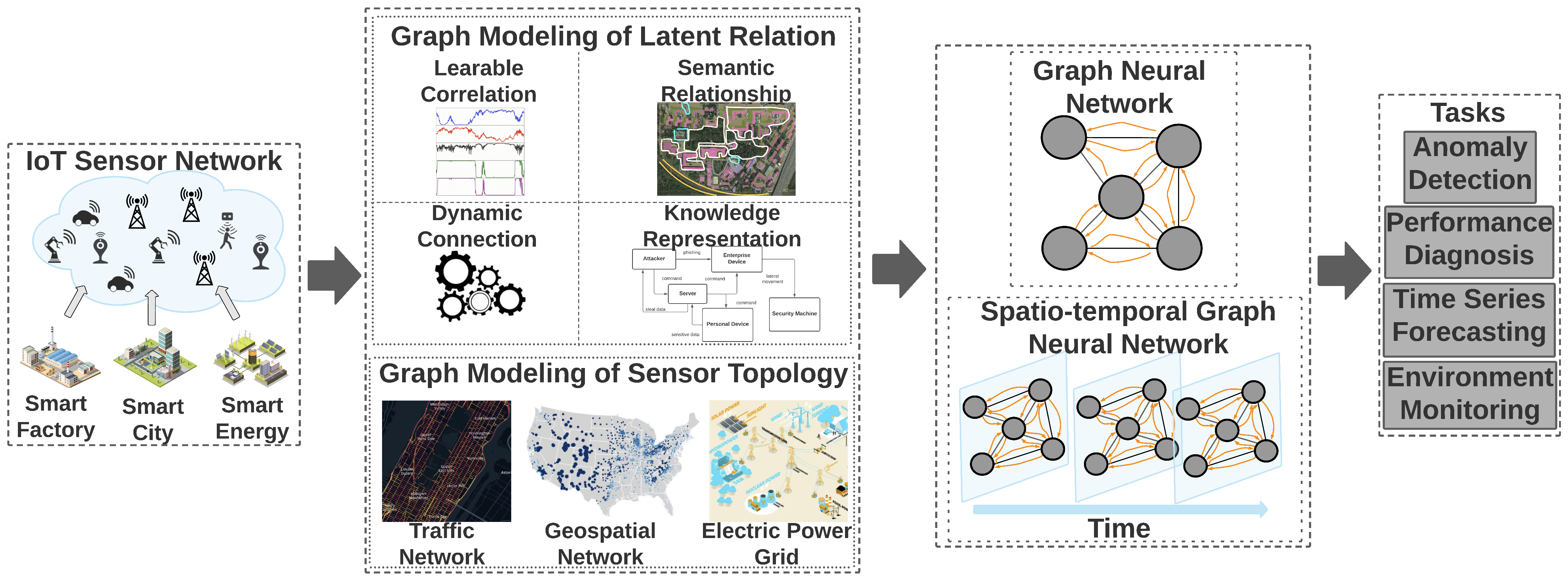}
\caption{General Graph Neural Network architecture to model complex Interconnection in IoT sensors}
\label{fig:gnn_sensor}
\end{figure}
\end{center}

\subsection{IoT Sensor Interconnection (ISI)}

Internet of Things (IoT) is the interconnection of digital objects and electronic devices in order to form an intelligent network among them and allow them to interact with each other. These interconnections among different types of sensors play tremendous part in our society and have been widely discussed in both transformative and industrial IoT scenarios. By leveraging building sensors (e.g., temperature, VOC, lighting), the interconnections between these sensors are able to help build a comfortable work place for smart buildings. Similarly, body sensors (e.g., smart bands and watches) can help athletes, fitness enthusiasts, and recreational sportspeople track energy expenditure levels, calories burned, miles traveled and more; chemical sensors (e.g., carbon dioxide, oxygen, electronic nose and catalytic bead) interact with environment sensors (e.g., optical, soil nutrient, airflow) to provide data that helps farmers monitor growth and optimize production of crops under harsh environmental conditions and challenges.  Different traffic sensors (e.g., environmental surveillance sensors, parking sensors, and speedometer sensors) are combined to build a robust traffic network. Smart grid utilizes a diverse set of sensors (e.g., thermal, temperature, humidity, smart meters) to ensure efficient generation, transmission, and distribution of power from its source to the end users. Smart factories use industrial sensors (e.g., pressure, proximity, fluid velocity, infrared meters) to enhance productivity and achieve industry 4.0. ISI has penetrated into almost every digital smart environment/system since a single sensor often fails to collect sufficient data to support most smart applications. Given that exposing critical latent information in the majority of intelligent applications requires collaborations from interconnected sensors, modeling IoT sensor data and utilizing the interconnections between different types of sensors plays a vital role in building smart cities. 

There is a wide range of research focusing on modeling IoT sensor data in the above mentioned smart application domains. In smart building applications, researchers pay close attention to areas such as energy-efficiency, privacy and security, and livable environment~\cite{qolomany2019leveraging}. Current methods rely on the development of cutting-edge machine learning and deep learning techniques to perform data-driven approaches to provide better solutions for the above cases. Somu et al.~\cite{somu2021deep} presented kCNN-LSTM, a deep learning framework that combines kmeans clustering, convolutional neural network, and long short term memory. They applied the proposed framework on energy consumption data recorded at predefined intervals to understand energy consumption patterns and provided accurate building energy consumption forecasts to achieve better energy-efficiency. Privacy and security are one of the biggest concerns in the modeling of sensor data. Gao et al.~\cite{gao2021decentralized} proposed PriResi to overcome the privacy issue when collaboratively training sensor data in residential buildings. PriResi combines cutting edge federated learning (FL) with LSTM, which allows residents to process all collected data locally while achieving collaborative training for load forecasting. Air quality in our living environment is critical to our health especially during the Covid-19 pandemic. Deep-Air ~\cite{han2021deep} introduced a novel hybrid CNN with LSTM to provide fine-grained city-wide air pollution estimation and station-wide forecasts on air quality. In smart industry, Wen et al.~\cite{wen2021remaining} developed a hybrid sensor fusion system to accurately predict the remaining useful life (RUL) of IoT-enabled complex industrial systems. These advanced learning algorithms have also been deployed to help in smart health, smart grid, and other domains~\cite{gao2020smartly, vimal2021iot, hasan2019electricity, li2017everything}. However, deep learning models such as those based on RNNs or CNNs fail to 1) explicitly leverage this potentially rich source of domain-knowledge into the learning procedure and 2) study the geospatial implications and latent connections among sensors.

GNNs set steps to overcome the above mentioned issues to better model IoT sensor data interconnections. GNNs take geospatial implications into account by combining the graph structured relational information to link the IoT sensors placed at different locations, and achieve better prediction performance for applications such as smart traffic network and power grid network. GNNs also help to find the implicit interconnections among sensors. For example, in a multi-modal sensor network (e.g., lighting, environment), each sensor can be represented as a node in a graph, and their latent interconnections need to be learned by using data-driven approaches. Established works \cite{zhang2019modeling, deng2021graph, lin2021multilabel, ouyang2021combining, liang2020deep, zou2020multi, narwariya2020graph, ding2021semi, almeida2021sensorimotor, fischer2021stickypillars, chen2020inference, wu2020connecting, bi2019graph, shrivastava2022graph, tekbiyik2021graph, boyaci2021graph, xu2021graph, lo2021graphsage, wu2021graphAD, zhang2020multi, chen2019gated, ouyang2021spatial, zhang2020semi} illustrated the performance of applied GNNs in smart city applications that involved IoT sensor interconnections. Table ~\ref{table:gnn_ISI} summarizes the sensor infrastructures, GNN models, and learning targets in the collected works. Despite numerous implementations of GNNs in different research tasks and domains, GNN modeling of IoT sensor interconnections can be generalized as the framework shown in Fig ~\ref{fig:gnn_sensor}. GNN modeling for IoT sensor interconnection consists of three parts: 1) sensors in IoT, 2) graph modeling of ISI, 3) graph neural networks for ISI.

\subsubsection{\textbf{Sensors in IoT}}
Smart IoT applications, such as anomaly detection\cite{wu2021graphAD, deng2021graph}, remote sensing\cite{liang2020deep}, and traffic prediction\cite{zhang2020semi}, require huge amount of sensor data to achieve diverse goals. In complex IoT sensing, different sensor infrastructures play important roles to capture various data representations from different angles. It is usually unrealistic for a single sensor to take geospatial information and sensor latent interconnections into consideration to model complex IoT problems, since various sets of sensors are required to coordinate with each other to provide a comprehensive understanding of the IoT environments.

IoT sensors can be classified based on different IoT tasks. The first type of sensors is environment monitor sensors, where the function is to collect time series data to monitor the changes of the monitored areas. Monitor sensors include temperature, pressure, volatile organic compounds (VOCs), $CO_{2}$, lighting, humidity, proximity, anisotropic magneto resistance (AMR), etc. Monitor sensors are mostly placed in smart buildings, grids and factories to capture the dynamic changes in real-time, which provides the opportunity for smart applications to continue using them in their best operations. The second type of sensors is surveillance sensors. They are widely used to take images or capture sensitive signals to detect and recognize surrounding objects, such as real-time traffic, building footprints, and human activities. Common surveillance sensors include camera, ultrasonic, GPS, satellite, aerial cameras, light detection and ranging (LiDAR), etc. Such sensors are often deployed in our cities to organize traffic, arrange parking spaces, and even monitor autonomous vehicles to create a smarter city. 
There are also other types of sensors we commonly use to serve other IoT applications, such as on body sensors like EGG~\cite{khan2020all}. Multiple EEG sensors can be used simultaneously in a cost-effective and straightforward way to assess bowel motility, gastric and cardiac activity for health monitoring. For smart agriculture, RFID sensor tags~\cite{finkenzeller2010rfid} detect environmental changes and events, and transmit the data wirelessly to RFID readers. These telemetry products are ideal in situations where measurement data need to be collected remotely and automatically. Depending on the types of RFID sensor tags, they can sense changes in motion, humidity, temperature, pressure, and more.

As shown in Table ~\ref{table:gnn_ISI}, the surveyed papers utilize different types of sensors to collect data for different smart applications. For researches on aerial image classifications, remote sensing and smart city applications like traffic prediction, authors used advanced computer vision techniques with a combination of aerial cameras and satellite, ultrasonic, GPS, and LiDAR to generate data. We discussed the advantages and disadvantages of different sensors in Section ~\ref{section4}. For research on smart buildings, factories, and grids applications like air quality prediction, anomaly detection, HVAC system, and remaining useful life estimation, authors need a combination of different types of monitor sensors to continuously collect time series data to keep track of the change in data. For example, studies in \cite{trinchero2011integration, song2012modeling, yongqing2013research, phala2016air} focused on temperature measurements to guarantee HVAC, and safety. Early fire detection tends to use environment monitor sensors such as temperature sensors (e.g., RTD, NTC thermistor, thermopile, thermocouple) and humidity sensors (e.g., capacitive, resistive) to sense the moisture and temperature change through the electric signals and give users an early notification. Home and factory gas leakage, commercial activities, and natural disaster are some other major problems that environment monitor sensors are deployed to help. Research in \cite{kumar2014energy, dong2017mems, gaur2019fire, kumar2011energy} applied the above mentioned temperature sensors together with smoke/gas sensors (e.g., $CO$, $CO_{2}$, $NO_{x}$, $PM_{x}$) as an indicator of fire which led to better application performance. Additional examples utilizing sensor interconnection include the combination of light/photon sensors and touch sensors. Light/photon sensors (e.g., infrared sensors, ultrasonic, proximity) are capable of sensing human movements or other objects within the range of 10-14 meters from the sensor. Touch sensors (e.g., wire resistive, surface capacitive, acoustic) are used to sense touching or near proximity. These sensors are often deployed together for applications such as outdoor lighting, parking space arrangement, building and shared space occupancy detection~\cite{aliyu2018towards, singh2016intelligent, tang2015augmented}.


\subsubsection{\textbf{Graph Modeling of IoT Sensor Interconnection}}

Graph modeling plays an important role in IoT sensor network, especially for sensor interconnections. In real-world IoT application, multi-modal and/or internet of sensors are implemented to capture dynamics of sensing objects. For example, $CO_{2}$ sensors in smart building can detect the level of $CO_{2}$ changing through time. However, it is impossible to know the cause of such changes with only the $CO_{2}$ data. Together with motion sensors, we can detect if it is caused by certain human activities that result in the change of $CO_{2}$ level. It is essential to understand the interconnections among different IoT sensors by observing their topological structure and capturing the correlations among the sensor data. In general, there are two kinds of graph modeling methods for ISI: 1) using graph to generate geospatial implications, such as modeling traffic network or power grid network; 2) using graph to model the implicit interconnections among sensors to find latent relationships, such as sensor inter-correlations within multi-modal sensor network.


\noindent\textbf{Node representation}: in graph modeling of ISI, we divide node representation into two parts as shown in Fig~\ref{fig:gnn_sensor}. First, in geospatial implication scenarios like traffic networks, geospatial networks, and power grids, node representation determines the geospatial meaning, that is the sensor as a node indicates its relative position in the city. For example, in traffic networks, the installed traffic camera and ultrasound sensors that continuously collecting data are the nodes in this case. In the papers that we surveyed, Zhang et al. \cite{zhang2020semi} proposed a novel semi-supervised hierarchical recurrent graph neural network (SHARE) for predicting city-wide parking availability by developing a graph structure from sensors such as camera, ultrasonic and GPS. Then they developed a context graph convolution block and a soft clustering graph convolution block to capture both local and global spatial dependencies between parking lots. Similar to this work, authors in the following studies \cite{zhang2020multi, wu2021graphAD, ouyang2021combining, kipf2016semi, boyaci2021graph, velivckovic2017graph} also created graph representations from geospatial sensors. They used the located cameras and satellite as nodes to create graph structures to perform traffic network prediction. They also used smart meters as nodes in the electricity grid graph and aerial cameras as nodes for remote sensing. 

Second, in implicit interconnection scenarios such as knowledge representation, dynamic connection, and latent learnable correlation, node representation determines semantic implication. For example, in a multi-modal sensor network each sensor can be represented as a node in a graph, but their latent interconnection needs to be learned by using data-driven approach. Zhang et al. \cite{zhang2019modeling} proposed a graph neural network-based modeling approach for IoT equipment (GNNM-IoT). By utilizing the environment monitor sensors such as temperature, pressure and voltage as nodes simultaneously, they considered temporal and intrinsic logical relationships of the data to perform air conditioning prediction with high accuracy. The result demonstrates better performances than those from ARIMA and LSTM methods. Other related works \cite{deng2021graph, shrivastava2022graph, chen2019gated, zou2020multi} also tried to find latent relationships among sensor data in the above mentioned scenarios by using the sensor to create a graph structure, followed by data-driven approaches to learn the correlations. For example, Chen et al. \cite{chen2019gated} worked on the traffic prediction problem, however, they used cars as nodes to create the graph structure, and proposed a novel end-to-end multiple Res-RGNNs framework to find the dynamic connections among cars while discovering latent relationships between cars before performing traffic prediction.  


\noindent\textbf{Edge connection}: edge connection means how the nodes in the graph structure connect and interact with each other. Based on the node representation, adding edges between two related sensors can 1) clearly state the geospatial connection with natural setting between sensors and 2) help learn implicit interconnections through data-driven approaches. In our surveyed papers, authors in \cite{zhang2020multi, wu2021graphAD, ouyang2021combining, kipf2016semi, boyaci2021graph, velivckovic2017graph} use sensor location as node representation, and geospatial connection as edge connection to build the graph structure.In geospatial implication scenarios where cameras are deployed in the city and smart meters are placed in the electricity grid in fixed locations, it is much easier and more accurate to model the graph using sensor topology based on physical sensor information. As for the implicit interconnection scenarios, edge connections can be learned through data-driven approaches to connect two sensors. Different from geospatial connection, the node representation in this case may or may not have natural connections. Edge connection between sensors can be their dynamic connections or knowledge representation. Works in \cite{chen2019gated, deng2021graph, shrivastava2022graph, zou2020multi, lin2021multilabel, bi2019graph, wu2021graphAD} used deep learning methods such as CNNs and RNNs to generate potential representations of the edges, which encode the temporal and local graph structural changes of the nodes in the graph. 


\subsubsection{\textbf{Graph Neural Networks for IoT Sensor Interconnection}}
The graph representation for ISI can create graph structured data that are either with geospatial connection or latent connection, which traditional deep learning models such as CNNs and RNNs can not provide. GNNs in ISI are developed to dig into the complex interactions between sensors and capture the correlations among data streams from the graph structure. There are two types of GNNs in our surveyed studies as shown in Table~\ref{table:gnn_ISI}. The first type of GNNs for ISI is spatio-temporal GNNs, which are made of static structures and time-varying features, and such information in a graph requires a neural network that can deal with time-varying features as demonstrated in the these works~\cite{zhang2020multi, wu2021graphAD, ouyang2021combining, kipf2016semi, boyaci2021graph, velivckovic2017graph, chen2019gated, narwariya2020graph, ding2021semi}. The second type of GNNs for ISI is graph autoencoder, which maps graph data into a low-dimensional space for graph analytics~\cite{wu2021graphAD, zhang2019modeling, deng2021graph, shrivastava2022graph}. 

\noindent\textbf{Spatio-temporal GNNs}: systems consisting of information about structural relationships in space and time can be considered as a spatio-temporal graph. These graphs allow a neural network to deal with static structures and time-varying features. Ouyang et al.\cite{ouyang2021spatial} proposed a Spatial-Temporal Dynamic Graph Convolution Neural Network (ST-DGCN). They developed a dynamic adjacency matrix in the graph convolution layer that extracts latent and time-varying spatial dependencies. In order to jointly model spatial and temporal dependencies, they combined dynamic graph convolution with gated recursive units and proposed a unified DGC-GRU for air quality prediction. Zhang et al. \cite{zhang2020semi} proposed a semi-supervised spatial-temporal learning framework that incorporates environmental contextual factors and sparse real-time parking data into environmental factors and sparse real-time parking availability data for city-wide parking availability prediction. Chen et al. \cite{chen2019gated} proposed to jointly capture graph-based spatial dependencies and temporal dynamics so that the spatio-temporal features of the traffic network can be extracted appropriately. Liu et al. \cite{liu2020st} proposed a spatio-temporal multi-modal fusion model (ST-MFM) and used GCN to extract the spatial features. Their approach can simultaneously learn latent graph structure and node feature representation for anomaly detection using traffic data.

 
\noindent\textbf{Graph autoencoders}: graph auto-encoder and its variants have been primarily used for representation learning on graph-structured data.  It aims to preserve the graph structure of the input matrix (e.g., adjacency matrix) in a low-dimensional space
by matrix factorization. Zhang et al. \cite{zhang2019modeling} proposed GNNM-IoT to model relationships between embedded sensors in IoT equipments. They used an encoder to design and learn the relationships between vertices. A decoder was then used
to reconstruct the input data based on the vertices and the relation function. Berg et al. \cite{berg2017graph}  proposed GC-MC, a graph-based auto-encoder framework for
matrix completion. The autoencoder generates potential features of user and item nodes through a form of information transfer on a bipartite interaction graph. These potential user and item representations are used to reconstruct the rating links by means of a bilinear decoder. The benefit of formulating matrix completion as a link prediction task on bipartite graphs becomes particularly apparent when the recommendation graph is accompanied by structured external information.

\begin{table*}[t]
    \centering
    \renewcommand{\arraystretch}{1.35}
  \resizebox{1\textwidth}{!}{
    \begin{tabular}{c|c|c|c|c}
    \toprule
   Sensors & Input Data
    & Models & GNN Kernels & Tasks  \\
     \midrule
      Temperature, Pressure, Voltage &  HVAC & GNNM-IoT\cite{zhang2019modeling} &MPNN\cite{scarselli2008graph}&  Air Conditioning Prediction

\\
      Water & SWaT,WADI Water Treatment & GDN\cite{deng2021graph}  &GDN\cite{deng2021graph} & Anomaly Detection
\\
      Satellite, Aerial Cameras & UCM Image & ML-CG\cite{lin2021multilabel} &GCN\cite{kipf2016semi}& Aerial Image Classification
\\
      Satellite, Aerial Cameras & UCM Image & AttResUNet\cite{ouyang2021combining} &GCN\cite{kipf2016semi}& Scene Classification
\\
      Satellite, Aerial Cameras & AID Image & CNN-GCN\cite{liang2020deep} &GCN\cite{kipf2016semi}& Scene Classification
\\
      3D depth cameras, Scanners & Point Cloud & MAUGNN\cite{zou2020multi} &MPNN\cite{scarselli2008graph}& Point Cloud Learning
\\
      Aircraft Turbofan Engine &  Multivariate Time Series & GNMR\cite{narwariya2020graph} &GAT\cite{velivckovic2017graph}& Useful Life Estimation

\\
      AVIRIS, Aerial Cameras & Pavia University, Salinas Image & DARMA-CAL\cite{ding2021semi} &GCN\cite{kipf2016semi}& Image Classification
\\
       Cable-actuated hands&Finger positional motion& Sensorimotor Graph\cite{almeida2021sensorimotor} &GFT\cite{shuman2013emerging}& Hand Dynamics Learning
\\
      LiDAR & Point Cloud KITTI & StickyPillars\cite{fischer2021stickypillars} &GCN\cite{kipf2016semi}& Point Cloud Registration
\\

      Ergodic theorems& Voter model Time Series& NC-GGN\cite{chen2020inference} &GCN\cite{kipf2016semi}& Network Inference
\\
      Camera& Traffic information & MTGNN\cite{wu2020connecting} &GCN\cite{kipf2016semi}& Time Series Forecasting
      
\\      
      NVS& N-MNIST, MNIST-DVS, N-CARS& RG-CNN\cite{bi2019graph} &GCN\cite{kipf2016semi}& Object classification
\\
      VOC, $CO_{2}$& Smart building IoT sensor network& DGCNN-GF\cite{shrivastava2022graph} &GCN\cite{kipf2016semi}& Prediction in Sensor Networks
\\
      LEO Satellite& SNR& RIS-GAT\cite{tekbiyik2021graph} &GAT\cite{velivckovic2017graph}& Channel Estimation
\\
      Smart Meters & Electricity Consumption& FDIAs-GNN\cite{boyaci2021graph} &GCN\cite{kipf2016semi}& Attack Detection
\\
      Smart Meters & Electricity Consumption& DHVG\cite{xu2021graph} &GCN\cite{kipf2016semi}& Anomaly Detection
\\
      Temperature, Water Stations& BoT-IoT, ToN-IoT& E-GraphSAGE\cite{lo2021graphsage} &GCN\cite{kipf2016semi}& Intrusion Detection
\\
      Variant of Industrial Sensors& U-turn, Taxi& Variant of GNNs\cite{wu2021graphAD} &GAT\cite{velivckovic2017graph}, GCN\cite{kipf2016semi}& Anomaly Detection
\\
      Camera, Satellite& Traffic Accident, Google Map Image& GraphCast\cite{zhang2020multi} &GAT\cite{velivckovic2017graph}& Traffic Risk Forecasting
\\
      Camera, Satellite& META-LA Traffic& MRes-RGNN\cite{chen2019gated} &GCN\cite{kipf2016semi}& Traffic Prediction
\\
      $PM_{2.5}$, $CO_{2}$& Air Quality Time Series& ST-DGCN\cite{ouyang2021spatial} &GCN\cite{kipf2016semi}& Air Quality Prediction
\\
      Camera, Ultrasonic, GPS& Baidu Map Time Series & SHARE\cite{zhang2020semi} &GCN\cite{kipf2016semi}& Parking Availability prediction
\\

   \bottomrule
\end{tabular}
}
\caption{Summary of Graph Neural Networks to IoT sensor interconnection}
\label{table:gnn_ISI}
\end{table*}

\section{Data and Application} \label{data_application}

\subsection{Public Datasets and Toolkits}

\begin{table*}[t]
    \centering
    \renewcommand{\arraystretch}{1.35}
  \resizebox{1\textwidth}{!}{
    \begin{tabular}{c|c|c|c|c}
    \toprule
       Dataset Name & Domain & Data Description & Relevant Studies & Link \\
     \midrule
     OpenAI Gym \cite{1606.01540} & Robotics & 
Gym is a toolkit for developing reinforcement learning algorithms. It can simulate continuous control tasks in robotics. & \cite{wang2018nervenet}  & \href{https://gym.openai.com/}{[url]} \\
    \midrule
    NGSIM US-101 \cite{us101} & \multirow{3}{*}{Autonomous Vehicle} & collected detailed vehicle trajectory data on southbound US 101 & \cite{jo2021vehicle} & \href{https://www.fhwa.dot.gov/publications/research/operations/07030/}{[url]}\\
    I-80 \cite{i80} & & collected detailed vehicle trajectory data on southbound US I-80 & \cite{jo2021vehicle} & \href{https://www.fhwa.dot.gov/publications/research/operations/06137/}{[url]}\\
    Stanford Drone \cite{robicquet2016learning} & & collects images and videos of various types of agents that navigate in a real world outdoor environment & \cite{dong2021multi} & \href{https://cvgl.stanford.edu/projects/uav_data/}{[url]} \\
    \midrule
     TST V2 \cite{h2qp48-16} & \multirow{2}{*}{Fall Detection} & The fall detection dataset contains depth frames, skeleton joints, acceleration & \cite{keskes2021vision} & \href{https://ieee-dataport.org/documents/tst-fall-detection-dataset-v2}{[url]}\\
     FallFree \cite{alzahrani2017fallfree} &  & Computer vision-based
applications pertinent to people who use a cane as a mobility aid & \cite{keskes2021vision} & Contact Author\\
    \midrule
NTU RGB+D \cite{shahroudy2016ntu} & \multirow{7}{*}{HAR} & Large-scale dataset for human action recognition  & \cite{zheng2019fall, keskes2021vision} &  \href{https://rose1.ntu.edu.sg/dataset/actionRecognition/}{[url]} \\
    MobiAct \cite{vavoulas2016mobiact} & & HAR dataset contains accelerometer, gyroscope, and orientation sensors data & \cite{mondal2020new} & \href{www.bmi.teicrete.gr}{[url]} \\
    WISDM \cite{kwapisz2011activity} & & HAR dataset collected labeled accelerometer data from daily activities & \cite{mondal2020new} & \href{https://archive.ics.uci.edu/ml/datasets/WISDM+Smartphone+and+Smartwatch+Activity+and+Biometrics+Dataset+}{[url]} \\
    MHEALTH \cite{banos2014mhealthdroid} & & HAR dataset contains acceleration, gyroscope, magnetic field, ecg data & \cite{mondal2020new, uddin2021human} & \href{http://archive.ics.uci.edu/ml/datasets/mhealth+dataset}{[url]}\\
    PAMAP2 \cite{reiss2012introducing} & & HAR dataset contains IMU hand, IMU chest, IMU ankle, and heart rate data & \cite{mondal2020new} & \href{https://archive.ics.uci.edu/ml/datasets/pamap2+physical+activity+monitoring}{[url]} \\
    HHAR \cite{stisen2015smart} & & HAR dataset contains accelerometer and gyroscope data & \cite{mondal2020new} & \href{http://archive.ics.uci.edu/ml/datasets/heterogeneity+activity+recognition}{[url]} \\
    USC-HAD \cite{zhang2012usc} & & Use off-the-shelf sensing platform called MotionNode \cite{motionnode} to capture human activity signals & \cite{mondal2020new} & \href{https://sipi.usc.edu/had/}{[url]}\\
    \midrule
     MASS-SS3 \cite{o2014montreal} & \multirow{2}{*}{Sleep Quality} & Collaborative database of laboratory-based polysomnography (PSG) recordings for sleep quality prediction & \cite{jia2020graphsleepnet} & \href{http://ceams-carsm.ca/mass/}{[url]}\\
     ISRUC \cite{khalighi2016isruc} & & a polysomnographic (PSG) dataset that allocate with sleep quality & N/A & \href{https://sleeptight.isr.uc.pt/}{[url]} \\
     \midrule
     Beijing \cite{beijingair} & \multirow{5}{*}{Air Quality} & Hourly scaled dataset of pollutants ($PM2.5, PM10, NO2, SO2, O3, CO$) from 76 station & \cite{qi2019hybrid, ge2021multi, wang2021modeling} & \href{https://beijingair.sinaapp.com/}{[url]}\\
     Tianjing \cite{urbancom} & & Features are same as Beijing Dataset & \cite{wang2021modeling} & \href{http://urban-computing.com/data/Data-1.zip}{[url]}\\
     CMA China\cite{zhang2017cautionary} & & Wind speed, wind direction, temperature, pressure and relative humidity for the same period & \cite{qi2019hybrid, seng2021spatiotemporal} & \href{https://archive.ics.uci.edu/ml/datasets/Beijing+Multi-Site+Air-Quality+Data}{[url]}\\
     NCEP \cite{nceweather} & & Multi-source observations from stations, satellites, and radars with physical atmospheric model & \cite{ge2021multi} & \href{https://www.ncdc.noaa.gov/data-access/model-data/model-dataset/global-data-assimilation-system-gdas}{[url]} \\
     KnowAir \cite{wang2020pm2} & & Features are same as Beijing and Tianjing dataset & \cite{wang2020pm2} & \href{https://github.com/shawnwang-tech/PM2.5-GNN}{[url]} \\
     \midrule
     USGS \cite{usgs} & \multirow{3}{*}{Water System} & River segments that
vary in length from 48 to 23,120 meters & \cite{jia2020physics} & \href{https://waterdata.usgs.gov/nwis}{[url]}\\
    Water Quality Portal \cite{wqdh} &  & WQP is the largest standardized water quality dataset & \cite{jia2020physics} & \href{http://www.waterqualitydata.us/}{[url]}\\
    Water Calibration Network \cite{ostfeld2012battle}& & Containing 388 nodes, 429 pipes, one reservoir, and seven tanks & \cite{rong2021graph} & \href{https://uknowledge.uky.edu/wdst_models/2/.}{[url]}  \\
    \midrule
    Spain \cite{ji2015evaluation} & \multirow{3}{*}{Soil} & 20 soil moisture stations from North-Western Spain, for the years 2016-2017 & \cite{vyas2020semi} & \href{https://disc.gsfc.nasa.gov/information/documents?title=Hydrology\%20Documentation}{[url]}\\
    Alabama &  & 8 soil moisture stations
from Alabama, USA during the year 2017 & \cite{vyas2020semi} & \href{https://www.wcc.nrcs.usda.gov/scan/}{[url]}\\
    Mississippi &  & 5 soil moisture stations from Mississippi, USA during the year 2017 & \cite{vyas2020semi} & \href{https://www.wcc.nrcs.usda.gov/scan/}{[url]}\\
    \midrule
    GNN4Traffic \cite{jiang2021graph} & \multirow{3}{*}{Transportation} & Repository for the collection of Graph Neural Network for Traffic Forecasting & \cite{jiang2021graph} & \href{https://github.com/jwwthu/GNN4Traffic}{[url]} \\
    TLC Trip \cite{tlc_trip} & & Origin-Destination demand taxi dataset, trip record data & \cite{moreira2013predicting, zhang2021dneat} & \href{https://www1.nyc.gov/site/tlc/about/tlc-trip-record-data.page}{[url]}\\
    Kaggle Taxi \cite{taxi_kaggle} & & Taxi Trajectories for all the 442 taxis running in the city of Porto, in Portugal & \cite{moreira2013predicting} & \href{https://www.kaggle.com/c/pkdd-15-predict-taxi-service-trajectory-i}{[url]}\\
    \midrule
    Pecan street \cite{dataport} & Misc & Minute-interval appliance-level customer electricity use from nearly 1,000 houses and apartments & N/A & \href{https://dataport.pecanstreet.org/}{[url]}\\

   \bottomrule
\end{tabular}
}
\caption{Open Datasets for Graph Neural Networks in IoT Sensing}
\label{table:publicdata}
\end{table*}

In this section, we summarize common open data used in our surveyed papers in Table \ref{table:publicdata}. Although there are many more public data set available in the IoT sensing field, we only included data that have already been shown to be or are potentially useful for GNN modeling. 

We categorize these data in Table \ref{table:publicdata} into three categories - {\bf 1) graph/network structured data} {\bf 2) non-graph sensory data} {\bf 3) extension data}. Graph/network structured data refers to those data that exhibit graph structures, such as the transpiration network data and air quality situation network. Non-graph sensory data refers to common sensing data that can be transformed into certain graph structures through some pre-processing steps, like human activity recognition (HAR) datasets, sleep quality PSG sensors, and origin-destination demand data. The extension data are used as node/edge features or augmented data to enrich the information in given tasks. For example, the CMA China dataset in air quality category provides wind speed, wind direction, temperature, and other weather related information to better model the air quality prediction tasks. 

\subsubsection{Graph/Network structured data}

\begin{itemize}

\item Air quality: most of the air quality datasets contain air quality data among 6 types of air pollutants including PM2.5, PM10, NO2, CO, SO2, and O3 \cite{beijingair, urbancom, wang2020pm2, nceweather}. The air quality data were collected among various stations in different regions. Therefore, the geographical network relations naturally form graph structures. 

\item Water system: similar to air quality network, the distribution of water sources (e.g., river segments, water calibration network) is a type of network. For \cite{usgs, wqdh}, the river segments were defined by the national geospatial fabric and used for the National Hydrologic Model as described in \cite{regan2018description}. For \cite{ostfeld2012battle}, it contains 388 nodes/water stations, 429 pipes, one reservoir, and seven tanks. 

\item Soil: the soil datasets take soil moisture stations as graph nodes. For instance, Spain dataset \cite{ji2015evaluation} consists of 20 soil moisture stations from North-Western Spain, for the years 2016-2017, with a temporal resolution of 15 days based on the availability of the features. The other two datasets, Alabama and Mississippi, share similar data structure with the Spain dataset.

\item Transportation network: there exist rich sets of data in the transportation area. According to the GNN4Traffic survey \cite{jiang2021graph}, graph-related datasets can be categorized as transportation network data (e.g., OpenStreetMap), traffic sensor data (e.g., METR-LA, PeMS-BAY \footnote{https://github.com/liyaguang/DCRNN}), GPS trajectory data, location-based service data, etc. More detailed information can be found in Section 5 of \cite{jiang2021graph}. 

\item Fall detection: the classic fall detection problem and datasets are more like a branch of human activity recognition. Datasets TST V2 \cite{h2qp48-16} and FallFree \cite{alzahrani2017fallfree} are two computer vision based datasets, which take skeleton joints as graph nodes, and acceleration as node attributes. 

\item Energy: to the best of our knowledge, there are no public datasets used in energy forecasting tasks with GNN-based methods. However, the graph nature of energy networks (e.g., photovoltaic (PV) systems \cite{karimi2021spatiotemporal}) shows a great potential for applying GNN-related models. 

\end{itemize}

\subsubsection{Non-graph Sensory Data}

\begin{itemize}

\item Autonomous vehicle: as we mentioned in Section \ref{fig:gnn_multiagent}, interactions of intelligent entities are important in GNN sensing domain. Existing approaches model the interactions between agents/vehicles as graph nodes\cite{lee2019joint, jo2021vehicle, gammelli2021graph, dong2021multi}. NGSIM US101\cite{us101}, and I-80\cite{i80} are datasets that record vehicle trajectory on two different highways. Stanford drone datasets\cite{robicquet2016learning} collected images and videos of various types of agents (not just pedestrians, but also bicyclists, skateboarders, cars, buses, and golf carts) that are useful to trajectory prediction.

\item Human activity recognition (HAR): HAR datasets usually contain multiple sensor data from different modalities. MobiAct, WISDM, MHEALTH, PAMAP2, HHAR, USC-HAD all contain accelerometer data. MobiAct, MHEALTH, HHAR, USC-HAD have gyroscope data. PAMAP2 contains more information like heart rate, and IMU sensor data in different parts of human body (e.g., hand, chest, and ankle). Mondal et al. \cite{mondal2020new} constructed HAR graph using non-overlap time windows. 

\item Sleep quality: information extracted from polysomnographic (PSG) sensor can be a golden standard for sleep quality monitoring. MASS-SS3 \cite{o2014montreal} and ISRUC \cite{khalighi2016isruc} are two PSG recording datasets for sleep quality prediction. Take MASS-SS3 as an example, each recording contains 20 EEG channels, 2 EOG channels, 3 EMG channels, and 1 ECG channel, and the authors in \cite{jia2020graphsleepnet} constructed sleep stage graphs after the feature extraction process. 

\item Origin-destination demand (OD demand): as a type of transportation problem \cite{tlc_trip, taxi_kaggle}, the OD problem could be treated as network/graph related problem. However, some works \cite{li2020graph, dapeng2021dynamic} suggested that naive transportation networks (i.e., zones/regions as nodes, demand as edges) do not contain relationships between OD pairs. Therefore, state-of-the-art works take OD pairs as nodes and the relations between the pairs as edges. 

\end{itemize}

\subsubsection{Extension Data}

The extension data sources serve as additional information/attributes for graph data. Researchers can add more domain/expert knowledge to existing graph data. For instance, CMA China \cite{zhang2017cautionary} provides weather information to aid air quality prediction. NCEP \cite{nceweather} presented multi-source observations not only from stations but also from radars with physical atmospheric models.

\subsection{GNN Applications in IoT}


\subsubsection{\textbf{Robotics and Autonomous Vehicle}}
\label{robotapplication}
The rapid increased complexity of real-world robotic applications increases the needs of advanced robot planning and control algorithms. Nowadays, deep learning based methods provide a set of feasible solutions to robotics and autonomous vehicle planning and control problems \cite{karoly2020deep}. GNNs, as a set of more advanced models that can leverage the interactions and relation information by message passing between nodes, have been widely applied in robotics and autonomous vehicle area, which involves clustering of nodes (e.g., joints, swarm). Specifically, in the robotics field, we can classify existing approaches into two categories - 1) single robot and 2) multi-robots planning/decentralized control. For 1), some of the current GNN solutions treat joints of robot as graph nodes, connections between joints as edges, and predict the robot's behaviors or movements \cite{wang2018nervenet, kim2021learning}. Other research treats task-relevant entities like objects or goals as the graph nodes, and combines GNN with reinforcement learning to learn the best policy to carry out the targeted tasks \cite{lin2021efficient}. GNNs can be found more commonly in 2), since multi-robots planning or decentralized control is naturally suitable for graph-related methods, where a graph can be constructed by the interactions between robots. Existing works assume that each robot has access to perceptions of the immediate surroundings and that it can communicate with neighbor robots. Therefore, the GNN applied here is to control the entire swarm to transmit, receive, and process these messages between neighbor robots in order to decide on actions \cite{khan2020, hu2021vgai, gama2021graph, chen2021decentralized, wang2021heterogeneous, li2020graph, tolstaya2020learning, li2021message, tolstaya2020multi}. Autonomous vehicles are a type of more advanced robots with rich sensors, and have the capability to drive autonomously and safely on the road. Recent approaches applied GNN on connected autonomous vehicles (CAV) network, utilized spatiotemporal relations between CAVs \cite{chen2021graph, gammelli2021graph}. In summary, GNNs can leverage the network structures in swarm and CAV networks communication and controls. For more information about multi-agent interactions between robots/vehicles, please refer to \ref{fig:gnn_multiagent}.  


\subsubsection{\textbf{Health Inference and Informatics}}
Health inference and informatics, also known as smart health, is an important part in machine learning and data science applications. 
Recently, GNN has a wide range of applications in the healthcare domain, but most of the research uses electronic health records (EHRs) instead of sensing \cite{ahmedt2021graph, choi2020learning, sun2020disease, zhu2019graph, liu2020heterogeneous}. However, sensing techniques are especially useful in healthcare monitoring and intervention due to their ubiquitous and unobtrusive attributes. For instance, Wang et al. \cite{wang2021personalized} comprehensively reviewed how mobile sensing technologies affect personal and population health. Some existing research used sensor clustering as nodes and modeled behaviors based on the input sensor graphs \cite{dong2021influenza, dong2021semi}. Other works directly used sensors as graph nodes, and modeled the interactions between each sensor pair as edges, have proven to be helpful in fluid intake monitoring \cite{tang2022using, tang2022sensor}, sleep stage detection \cite{jia2020graphsleepnet}, and structural health monitoring \cite{bloemheuvel2021computational}. Apart from traditional healthcare problems, human activities are also correlated with human health state. For example, fall detection is critical in preventing elderly people from life-threatening situations. Existing research used skeleton joints as nodes to detect falls \cite{zheng2019fall, keskes2021vision}. In addition, general human activity recognition (HAR) with wearable/smartphone sensors by using GNN and activity graphs \cite{mondal2020new, uddin2021human} achieved state-of-the-art results, and could serve as a more comfortable and non-invasive solution. Overall, there is little research in the healthcare domain that utilizes IoT solutions and GNN methods. It may be because healthcare does not have natural graph scenarios like other fields (e.g., transportation network, robotics swarms). Therefore, further research on how to construct better graph scenarios for healthcare related problems is in high demand to get more reasonable and effective results. 


\subsubsection{\textbf{Environment}}
With the rapid development of environmental analytical tools and monitoring technologies, more and more data are being generated daily in the ﬁeld of environmental science and engineering (ESE). The new demand of having more advanced and powerful data analytical approaches have become more urgent. Traditional machine learning methods addressed many complex ESE problems\cite{zhong2021machine}, for example, like prediction of particulate matter(PM 2.5), water resource availability, and endocrine disrupting chemicals (EDCs), etc. The recent advancements in GNN better address these existing problems in the ESE field. {\bf Air quality}, there are many state-of-art works aim to predict the PM2.5/air quality using ST-GNN (GNN + RNN)\cite{qi2019hybrid, gao2021graph, xu2021highair, seng2021spatiotemporal,ge2021multi, ouyang2021spatial,wang2021modeling, wang2020pm2, liu2021new}, the rationale for using this model structure is because the air quality problem is naturally related to the regions (spatial-GNN) and changes over time (temporal-RNN). {\bf Water state}, in water state prediction and management field, most works also follow the same graph (regions as nodes) structure and focusing on problems like water flow prediction\cite{jia2020physics}, water quality prediction\cite{xu2021surface}, large sample hydrology problem \cite{sun2021explore}, and water network partitioning problem\cite{rong2021graph}, where \cite{xu2021surface, rong2021graph} only consider GNN rather than ST-GNN. {\bf Soil/Sediment}, fewer research is focused on soil related problems. Vyas et al.\cite{vyas2020semi} proposed a semi-supervised soil moisture prediction framework by using self-attention based temporal GNN, using both remote satellite and weather data to predict the soil moisture by locations as nodes. {\bf Public health}, population health statistics is critical to public health policy 
rulemaking, and it is an important task for ESE. Nguyen et al. \cite{nguyen2019estimating} estimates the county health indices by GNN,  and  construct the graph representation for each county by using the interactions between health-related features in the community. All the previous work indicates GNN has the potential power for ESE related problems. 


\subsubsection{\textbf{Transportation}}

Transportation systems play an important role in life nowadays. With urbanization and population growth, transportation systems are becoming increasingly complex. Transportation systems create incredible amounts of data by modern-day sensors like camera, GPS, loop detectors. Therefore, the idea of intelligent transportation systems (ITS) has been emerged in order to provide data-driven paradigms for more accurate traffic predictions or advanced control policies, which accelerated the development of many specific human-centered transportation scenarios. According to existing literature\cite{veres2019deep}, the intelligent transportation has three major parts: 1) Destination and demand prediction 2) Traffic state prediction 3) Combinatorial optimization. Although traditional machine learning methods \cite{yuan2019machine} and reinforcement learning methods \cite{haydari2020deep} already provide acceptable solutions for many ITS problems, due to the graph natural of traffic and road network, GNN-based models have demonstrated superior performance to previous approaches on ITS tasks \cite{jiang2021graph}. We structure this section followed by the type of problem being investigated. {\bf Destination and demand prediction}, are two types of trip prediction. Destination prediction focuses on where a vehicle or person will end and demand prediction is learning where and when a trip will start. Most works combine these two tasks as origin-destination(OD) demand prediction task. As ride-hailing services (Uber, Lyft) have become increasingly popular, predicting passenger demand is critical for ride-hailing service platforms to dispatch orders more effectively. In order to predict demand more accurately, existing approaches leverage massive spatial-temporal transportation network data and apply GNN-based prediction method to forecast the OD demand in advance. Zhang et al. \cite{zhang2021dneat} designed a dynamic auto-structural GNN model to predict the OD demand using transportation graph where regions as nodes and OD pairs as edges. Li et al.\cite{li2020graph} proposed a probabilistic graph convolution model (PGCM) for OD demand prediction and demand interval approximation. The model use OD-pairs as graph node, treat the correlation/similarity between the demand patterns of two OD pairs as edges, where the an OD-pair indicates one region (the origin) to another region (the destination). They claim the naive transportation network (i.e., zones/regions as nodes, demand as edges) will not contain the information between OD pairs, which is essential for OD demand estimation. Zhang et al. came out with another paper \cite{dapeng2021dynamic} following the same graph design as \cite{li2020graph}, and presented a novel dynamic node-edge attention network to address the challenges from the demand generation and attraction perspectives. The proposed solution captures the temporal evolution of node topologies in dynamic OD graphs and evaluates the model on both China(DiDi Chengdu) and USA\cite{tlc_trip} real-world ride-hailing demand datasets. Other than ride-hailing services demand, bike sharing system becomes extremely popular especially in China. The demand of maintain bike availability is essential for bike-sharing system operators. Zhu et al. \cite{zhu2022redpacketbike} presented a context-aware deep neural network simultaneously modeling the spatial, temporal, contextual information by GCN, RNN and Autoencoder. Their model provided optimal solutions to rebalance city-wide stations by integer linear programming (ILP). All the previous works indicate GNN has the capability to capture spatial or relation connections between region nodes or OD-pair nodes in demand prediction problems. {\bf Traffic state prediction}, traffic state prediction including traffic forecasting, traffic flow prediction, and travel time estimation, etc. This problem is important and well studied in the transportation field. Many existing surveys addressed the traffic prediction problem through deep learning methods to graph neural network methods\cite{chen2021survey, jiang2021dl, yin2021deep, jiang2021graph, yuan2021survey, nagy2018survey}. Due to the network nature of traffic flows, GNN fits the mold to model the interaction between placed sensors via GPS location or roads \cite{liu2021traffic, xia20213dgcn, jin2022stgnn}. For more details of using GNN to forecast traffic, we refer to \cite{jiang2021graph}. {\bf Combinatorial optimization},  is an important set of transportation problems attached to the domain of combinatorial optimization. These problems have extremely high dimensional solution space (NP-hard to solve) and are often raised in navigation or traffic management systems \cite{veres2019deep}. One of the most famous example of these types of problem is the traveling salesman problem (TSP), where a "salesman" visits a number of cities to sell their products, but they are only allowed to visit each city once and must return to the starting city at the end of the trip. Vehicle routing problems (VRP) are also a popular variant of the TSP problem, with solutions to VRP in high demanded among navigation companies. Cappart et al. \cite{cappart2021combinatorial} presented a conceptual review of how the state-of-art works use GNN to solve combinatorial optimization problems. Specific to transportation applications, Lei et al. \cite{lei2021solve} proposed a end-to-end residual edge-graph attention neural network which significantly improved the solution quality of TSP and VRP. 


\subsubsection{\textbf{Misc: Manufacturing, Energy, Smart Home}}
GNNs have also been widely applied in Misc fields, that is Manufacturing, Energy, and Smart Home. We present the following real-world applications as examples. Mozaffar et al. \cite{mozaffar2021geometry} applied GNN in additive manufacturing (AM). AM is commonly used to produce builds with comprehensive geometric. Current data-driven methods demonstrate their strong capability on AM processes. However, the generalizability of creating wide range geometry shapes remains challenging. Therefore, they use GNN to capture the spatio-temporal dependencies of thermal responses in AM processes, and the results show the high performance on long thermal histories of unseen geometries. Root cause analysis is also a challenge in manufacturing, specifically in multistage assembly lines. Leonhardt et al. \cite{leonhardt2021pen} proposed a GCN-based process estimator neural network (PEN), which solve the current limitation of linear approaches, that is the current solutions cannot process dense 3D point cloud data of the product. For energy fields, a larger number of photovoltaic (PV) systems have been added to electronic grids. Therefore, it's critical to predict the performance of PV and guarantee the reliability of PV systems. Karimi et al. \cite{karimi2021spatiotemporal} presented a spatiotemporal GNN model by utilizing the relationship between PV plants as graph to better forecast the PV power. For smart homes, many existing methods contribute to enhance a home's energy usage, efficiency, convenience, productivity, and life-qualities \cite{mohammadi2018deep}. This can be highly related to the robots section \ref{robotapplication}. Kapelyukh et al. \cite{kapelyukh2021my} train a variational autoencoder empowered GNN network to help robots to arrange household objects, the model take objects in a scene as nodes, and use a fully connected graph to predict the positions of objects. The above research shows GNN can help improve Misc in many aspects. 


\section{Challenges and Future Directions} \label{challenge_future}
\subsection{Reliability}
One of the critical requirements for the deployment of neural network models in the IoT is reliability, which aims to guarantee their functionality in real-world situations. Unsuccessful IoT communication, which can be caused by sensor malfunction, Internet interruption, and data corruption, can negatively impact the effectiveness of IoT applications with inaccurate or incorrect decision-making \cite{moore2020iot}. For example, in a smart city application of IoT, network infrastructure failure is fatal to the traffic management, which relies on processing the real-time update of sensor and traffic lights to control traffic flow. Additionally, in a smart health application, sensor and/or network failures might cause healthcare practitioners to miss the optimal treatment period, resulting in severe physical or mental damage to patients.
Thus, in spite of the superior performance that has been achieved by using GNNs, there is still concern in how failures can be prevented and detected as well as how the adverse exceptions can be addressed. Developing reliable neural network models is a promising research direction to realize large-scale IoT deployments. Liu et al. \cite{liu2020handling} propose a novel algorithm tying to alleviate sensor failure issue: by leveraging message passing mechanism in GNNs, the proposed algorithm can use the information passed from available sensor to reconstruct the features from missing sensors. However, since IoT systems always have a large number of sensors, transmitting all sensor readings to cloud for centralized data processing, prediction and controlling results in significant delays by communication network congestion, and computational overload \cite{al2020survey}. Therefore, efficient and effective edge computing can be investigated to processing the data at the individual sensor level or local sensor network areas. How to implement GNNs on edge computing systems is an exciting area of study to enhance reliability of IoT applications.

\subsection{Privacy Preservation}
IoT systems are laying the groundwork to link every object in our daily life together, transforming the world. Things will be able to connect and interact with one another and with their surrounding environment, aiming to improve people's quality of life. In the process of sensing the environment by the connected Things, information can be communicated with other IoT devices. The current strategy to store the sensory data is to upload the data to a central server (cloud servers) for future use. Despite the benefits of cloud-based data storage, centralized IoT systems may confront significant challenges. For instance, unencrypted server data is vulnerable to hacking and may result in the disclosure of critical information. Another example is, during the data collection in mobile sensing systems, GPS trajectories may expose the location of mobile users, which can be used to infer additional sensitive personal information such as race, gender, physical activity, social relationships, and health condition \cite{kargl2019privacy}. There are several privacy preservation techniques that can be utilized to enhance privacy protection in IoT systems, such as anonymization, federated learning, and differential privacy. For example, the k-anonymity method and variations have been developed to anonymize the identity of data items' sources \cite{huang2010preserving}. And differential privacy can minimize the disclosure of sensitive information by introducing perturbations that follow a specified noise distribution while preserving the value of the data \cite{liu2017deeprotect}. Federated learning (FL) is another prospective techniques for privacy preservation, where IoT nodes can collaboratively contribute knowledge to the global learning objective without transmitting data samples to a central server or exchanging data across IoT nodes. However, there exist disadvantages in anonymization techniques and differential privacy: anonymization exposes sensitive characteristics to inference attacks, while differential privacy reduces the usefulness of data and diminishes the strength of learnt models \cite{yin2021comprehensive}. Federated GNNs, which retain the predictive potential of GNNs while also protecting sensitive data, is an additional research path worth investigating. 

\subsection{Security}

The interconnected things in IoT introduce multiple security challenges, as the vast majority of Internet technologies and communication protocols were not created with IoT in mind \cite{mosenia2016comprehensive}. With the advancement of sensor technology and widespread use of IoT, consumers are concerned about their personal information, hidden cyber threats and technological crime, as well as other security problems that might constrain IoT applications. Massive amounts of data are collected and manged by IoT systems to support the services in healthcare, intelligent transportation, and smart agriculture, etc. Attackers and adversaries, such as cybercriminals, hackers, and hacktivists, want to profit from the priceless information. When IoT devices are hacked, credit card numbers, bank account PINs, and other personally identifiable information may be exposed to the public, resulting in financial loss for users of IoT services. IoT nodes and other IoT components that are connected through the Internet are also vulnerable to intelligent malware, which are designed to spread from device to device, interfering users' everyday life. What's more, hacktivists or illegitimate political agitators can exploit and manipulate the connected smart devices to spark protests against sovereign governments \cite{mosenia2016comprehensive}. There are already established security protection mechanisms in use, such as encryption, authentication, and access control. However these pre-defined mechanisms are incapable of adapting to new types of attacks and intelligent adversaries to provide adequately safeguard for IoT networks \cite{al2020survey}. Machine learning models (e.g. neural networks) have been extensively investigated to learn normal and abnormal patterns in the interactions of smart devices in IoT networks. From data-driven approaches, the information relating to the robust operations of IoT systems can be collected and utilized to profile normal interaction behaviors of IoT nodes; hence, malicious activities can be detected if the collected data shows anomalous distributions. However, existing methods of using conventional machine learning and neural network models cannot capture the topological information and multi-hop interactions between IoT nodes. There is an emerging trend of applying GNNs in IoT intrusion detection: for example, Lo et al. \cite{lo2021graphsage} propose a GNN-based network intrusion detection system, which can capture both the edge features
of an IoT network and the topological features. Originating in game theory, adversarial learning and GNNs have potential to be integrated in the future to learn malicious activities with economic consideration.

\subsection{Computational and Energy Efficiency}

Massive data production rates in the era of Big Data and the Internet of Things (IoT) continuously increase the demand for massive data processing, storage, and transmission \cite{shafique2017adaptive}. However, IoT devices are resource-constrained devices, with limited memory, computation and energy \cite{wang2020convergence}. IoT systems that are limited in terms of energy and power must not only provide high performance capabilities, but also serve as a platform for automation and intelligence. From handwritten digit recognition to autonomous driving, deep learning techniques (e.g., CNNs, RNNs) demonstrate groundbreaking success in achieving human-level recognition capability \cite{kuutti2020survey}. This superior performance has resulted in a huge increase in the use of DNN models in various Internet of Things (IoT) applications. However, the massive computational, energy, and storage requirements of DNN models make their implementation on resource-constrained IoT devices prohibitively expensive \cite{li2021block}.
Current solutions that implement DNNs in IoT devices to alleviate the issues of high demanding computational and energy resources include: 1) cloud computing; 2) developing edge computing GPUs. However, cloud computing suffers from high wireless energy overhead and has unattainable performance in weak network connectivity. GPUs in mobile devices may deplete significant amounts of mobile battery capacity \cite{lane2016deepx}. Different from the infrastructure improvement solution, DNN compression techniques have been proposed for efficient storage and computation consumption with minimal accuracy compromise \cite{mishra2020survey}. For example, knowledge distillation enables one to train a student model by learning from a teacher model, and the student model acts as a compressed version of the teacher model with similar prediction power \cite{sun2019patient}. The research of using knowledge distillation on graph neural networks has been investigated in these studies \cite{liu2019knowledge,lee2019graph, lassance2020deep}, while not extensively studied in the filed of IoT. In the future, there will be opportunities to research how to efficiently integrate GNNs in IoT devices by using knowledge distillation techniques. 

\section{Conclusion} \label{conclusion}
The usage of IoT devices has exploded in popularity over the past decade, owing to their ability to connect, cooperate, and collaborate on things in a plethora of domains of people's everyday lives. In the sensor-rich environment, conventional deep learning techniques have demonstrated their capacity to process complex multi-modal sensory data and sophisticated detection and recognition tasks. By considering IoT sensors' topology, Graph Neural Networks (GNNs) can capture the complex relationships and interdependency within sensor networks and demonstrate state-of-the-art results in numerous IoT tasks. In this article, we provide a comprehensive overview of the implementation of GNNs in the IoT field. Based on graph representation of IoT sensing, we propose a categorization scheme to classify the existing works of GNNs in IoT into graph modeling of multi-agent interaction, human behavior dynamics, and IoT sensor interconnection. For each category, we briefly summarize the sensor infrastructures, demonstrate the general GNN modeling frameworks by using the corresponding sensory data, and provide a detailed review of the graph components (e.g., nodes, edges) and the diverse GNN architectures. We present a review of public datasets that can be used for GNNs modeling in IoT. Additionally, we discuss the impact of GNNs and their practical implementation in IoT. Finally, we illustrate current research issues and directions for future study in GNN for IoT.


\bibliographystyle{ACM-Reference-Format}
\bibliography{reference}

\end{document}